\documentclass[10pt,twocolumn,letterpaper]{article}

\usepackage{cvpr}              % To produce the CAMERA-READY version
\usepackage{graphicx}
\usepackage{amsmath}
\usepackage{amssymb}
\usepackage{booktabs}
\usepackage{comment}
\usepackage[accsupp]{axessibility}  % Improves PDF readability for those with disabilities.

\usepackage{xcolor}
\newcommand{\cococolor}[1]{\textcolor[HTML]{1ad3f3}{#1}}

\usepackage[pagebackref,breaklinks,colorlinks]{hyperref}

\newcommand{\model}{\textsc{SmallCap}\xspace}

\usepackage{soul}

\usepackage[capitalize]{cleveref}
\crefname{section}{Sec.}{Secs.}
\Crefname{section}{Section}{Sections}
\Crefname{table}{Table}{Tables}
\crefname{table}{Tab.}{Tabs.}

\def\cvprPaperID{9761} % *** Enter the CVPR Paper ID here
\def\confName{CVPR}
\def\confYear{2022}

\begin{document}

\title{\textsc{SmallCap}: Lightweight Image Captioning Prompted \\with Retrieval Augmentation
}

\author{Rita Ramos\\
Institution1\\
Institution1 address\\
{\tt\small firstauthor@i1.org}
\and
Bruno\\
Institution1\\
Institution1 address\\
\and
Desmond Elliott\\
Institution1\\
Institution1 address\\
\and
Yova Kementchedjhieva \\
Institution1\\
Institution1 address\\
}

\author{Rita Ramos$^{\dagger}$ \ \ Bruno Martins$^{\dagger}$ \ \ Desmond Elliott$^{\star,\ddagger}$ \ \ Yova Kementchedjhieva$^{\star}$ \\
        $^{\dagger}$INESC-ID, Instituto Superior Técnico, University of Lisbon \\ $^{\star}$Department of Computer Science, University of Copenhagen \\ $^{\ddagger}$Pioneer Center for AI \\
        \texttt{ritaparadaramos@tecnico.ulisboa.pt}}
\maketitle

% If the title and author information does not fit in the area allocated, uncomment the following
%
%\setlength\titlebox{<dim>}
%
% and set <dim> to something 5cm or larger.

% Author information can be set in various styles:
% For several authors from the same institution:
% \author{Author 1 \and ... \and Author n \\
%         Address line \\ ... \\ Address line}
% if the names do not fit well on one line use
%         Author 1 \\ {\bf Author 2} \\ ... \\ {\bf Author n} \\
% For authors from different institutions:
% \author{Author 1 \\ Address line \\  ... \\ Address line
%         \And  ... \And
%         Author n \\ Address line \\ ... \\ Address line}
% To start a seperate ``row'' of authors use \AND, as in
% \author{Author 1 \\ Address line \\  ... \\ Address line
%         \AND
%         Author 2 \\ Address line \\ ... \\ Address line \And
%         Author 3 \\ Address line \\ ... \\ Address line}

%Rita Ramos
%INESC-ID, Instituto Superior Técnico, University of Lisbon

\begin{abstract}
  Recent advances in image captioning have focused on scaling the data and model size, 
  substantially increasing the cost of pre-training and finetuning.
  As an alternative to large models, we present \textsc{SmallCap}, which generates a caption conditioned on an input image and related captions retrieved from a datastore. 
  Our model is lightweight and fast to train, as the only learned parameters are in newly introduced cross-attention layers between a pre-trained CLIP encoder and GPT-2 decoder. 
  \model can transfer to new domains without additional finetuning and can exploit large-scale data in a training-free fashion since the contents of the datastore can be readily replaced. 
  Our experiments show that \textsc{SmallCap}, trained only on COCO, has competitive performance on this benchmark, and also transfers to other domains without retraining, solely through retrieval from target-domain data. Further improvement is achieved through the training-free exploitation of diverse human-labeled and web data, which proves to be effective for a range of domains, including the \texttt{nocaps} benchmark, designed to test generalization to unseen visual concepts.\footnote{Code: \url{https://github.com/RitaRamo/smallcap}.} 
\end{abstract}

% what is the external knowledge that the model can access - in the intro

% external data

\section{Introduction}
%Image captioning is a key task in multimodal learning wherein a caption is automatically generated to describe the visual content of an image.
% Problem statement

% \begin{figure}
%      \centering
%      \begin{subfigure}[b]{\linewidth}
%          \centering
%          \includegraphics[width=\textwidth]{images/main_model_comparison_coco.pdf}
%          \label{fig:y equals x}
%      \end{subfigure}
%      \hfill
%      \begin{subfigure}[b]{\linewidth}
%          \centering
%          \includegraphics[width=\linewidth]{images/main_model_comparison_nocaps.pdf}
%          \label{fig:three sin x}
%      \end{subfigure}
%         \caption{\model compared to other approaches in terms of number of trainable parameters and CIDEr score on the COCO dataset and the out-of-domain split of the \texttt{nocaps} challenge dataset. \model is competitive to other lightweight models on COCO, and outperforms much larger models on \texttt{nocaps}.}
%         \label{fig:my_label}
% \end{figure}

\begin{figure}
    \centering
    \includegraphics[width=\linewidth]{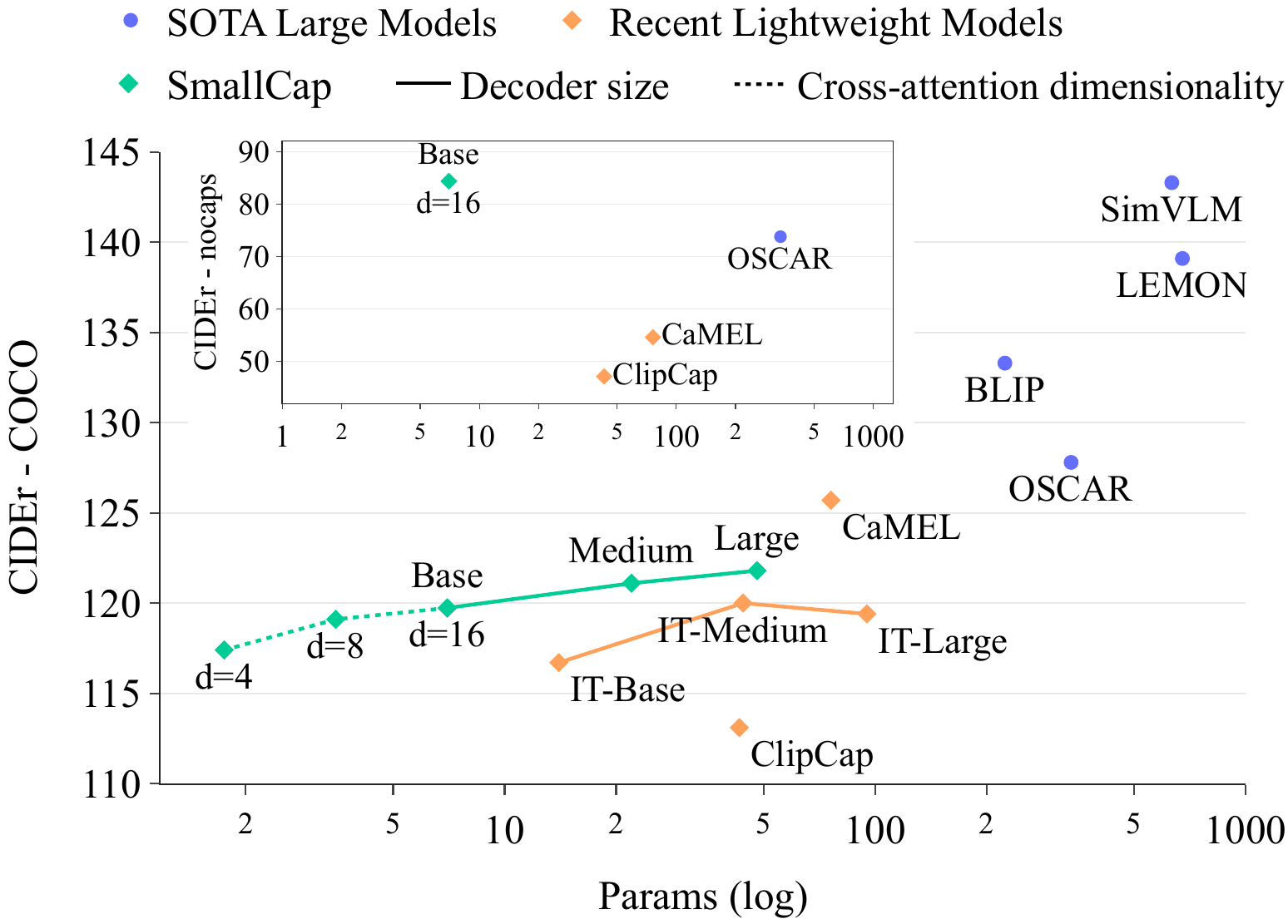}
    \caption{\model's performance on the COCO dataset and on the out-of-domain split of the \texttt{nocaps} dataset, compared to other approaches in terms of number of trainable parameters. We can control the number of trainable parameters through the dimensionality of the cross-attention  ($d=d_v=d_k$) and the size of the decoder. \model is competitive to other lightweight models on COCO, and outperforms much larger models on \texttt{nocaps}.}
    \label{fig:my_label}
\end{figure}

The state-of-the-art in image captioning
%, published by companies like Google, Microsoft and SalesForce, --> leaving this out as per Des's comment
is defined by increasingly large-scale models trained on increasingly large-scale datasets \cite{wang2021simvlm, hu2022scaling, wang2022git, li2022blip}. 
%This leads to high scores on the established performance metrics at the cost of extensive and expensive training.
Scaling up 
%favours industry labs~\cite{strubell-etal-2019-energy}
leads to higher computational demands for model pre-training and finetuning on downstream tasks. This becomes especially relevant when numerous model versions may be needed for different visual domains \cite{Agrawal_2019} and end-users in practical applications, e.g.\ image captioning for the visually impaired~\cite{gurari2020captioning}.
%This trend of scaling up hinders research as academia budgets fall short of the increasing compute and data demands \cite{strubell-etal-2019-energy}. 
%It also hardly aligns with the strive for a greener AI \cite{strubell-etal-2019-energy}: the promise of model reusability falls through when the growing size of pre-trained models translates into growing compute demands for finetuning them. The latter become especially relevant in real-world applications where numerous model versions may be needed for different end-users, use-cases and time periods.

%Model finetuning is task and data specific, which ramps up training costs 
Some efforts have been made recently to reduce the cost of model training, e.g., ClipCap \cite{clipcap} and I-Tuning \cite{luo2022tuning}. % present lightweight captioning models %, ClipCap and I-Tuning, respectively, 
These models use an off-the-shelf pre-trained vision encoder
%(CLIP \cite{}) 
and language decoder.
%(GPT2 \cite{}) 
The parameters of these pre-trained components are frozen and only a mapping between the two is trained for the task of image captioning. This results in a highly reduced number of trainable parameters ($\sim$43M in each case) and faster training time.
While these models operate on a much more manageable scale from a research perspective, they can still be unsuitable for the aforementioned practical applications, as both models require separate training for every use-case.

% Our contribution
%Des
This work presents \textsc{SmallCap}, an image captioning model, prompted with captions retrieved from an external datastore of text, based on the input image. This formulation of image captioning enables a range of desirable properties: lightweight training, training-free domain transfer, and exploitation of large data in a training-free fashion.  
%In this work, we present \textsc{SmallCap}, a captioning model which is augmented with retrieval, wherein knowledge from an external text source is presented to the model as a prompt in the generation process. This retrieval augmentation enables a range of desirable model features: lightweight training, training-free domain transfer, and exploitation of large data in a training-free fashion. 

\textsc{SmallCap} is both light to train and highly effective (see Figure~\ref{fig:my_label}).\footnote{The \texttt{nocaps} results shown in the figure include only models that follow the challenge guidelines, by training on the COCO dataset only.}
%This allows it to operate on a highly reduced budget of trainable parameters. 
It uses a pre-trained CLIP vision encoder \cite{radford2021learning} and GPT-2 language model \cite{radford2019language}, which are frozen and linked through new cross-attention layers, amounting to 7 million trainable parameters.
Through retrieval, the model leverages external data and therefore has to store less information within its 
weights (as demonstrated in Figure~\ref{fig:decay_cross_attention}). %which are trained only on COCO for 4 epochs. %We find that \textsc{SmallCap} performs on par with other lightweight models on the common COCO benchmark \cite{chen2015microsoft} despite an 83\% reduction in trainable parameters.
Trained on the common COCO benchmark \cite{chen2015microsoft}, \textsc{SmallCap} performs on par with other lightweight-training models, despite an 83\% reduction in number of trainable parameters. 

\textsc{SmallCap} can also leverage data in a training-free manner. Once the model is trained, we can replace the datastore with either (i) captions from a new domain or (ii) a large and diverse collection of captions. In the first case, which presents an alternative to finetuning, \textsc{SmallCap} gains access to the style and concepts that characterize the new domain and can generate captions accordingly. In the second case, which presents an alternative to generalized pre-training, \textsc{SmallCap} gains access to general knowledge that it can apply to any domain. Our experiments show that \textsc{SmallCap} effectively leverages new knowledge accessed through a retrieval-based prompt, improving its performance on different datasets. This includes the challenging VizWiz dataset, where images are captioned for the visually impaired \cite{gurari2020captioning}, and the \texttt{nocaps} challenge dataset with rarely-seen and unseen visual concepts \cite{Agrawal_2019}.

%\textsc{SmallCap} has only \textbf{29M} trainable parameters, which are trained \textbf{once} for \textbf{4} epochs on a \textbf{single} GPU card. 
\model competes with other lightweight-training models on in-domain evaluations and outperforms them by a large margin out-of-domain. It overcomes a key limitation of previous models, which require explicit finetuning to adapt to new domains, and in this way attests to the potential of retrieval augmentation for multimodal tasks. 
%\model competes with other lightweight models on in-domain evaluations and outperforms them by a large margin out-of-domain. Through retrieval, \textsc{SmallCap} overcomes a key limitation of previous lightweight models, which require explicit finetuning to adapt to new domains. 

\section{Related Work}

%[We need a sentence or two here introducing the section]

\subsection{Image Captioning Models}

 Current approaches to image captioning employ  encoder-decoder methods, where an input image is passed to a visual encoder and a caption is generated by an auto-regressive language decoder \cite{xu2015show, anderson2018bottom,yang2022reformer}. 
 %The vision encoder typically relies on an auxiliary object detector (e.g. a FasterRCNN model \cite{}), but recently an alternative approach, based on CLIP \cite{}, has been successfully applied in image captioning as well \cite{}. CLIP is trained with a vision-and-language contrastive objective of mapping images and text into a shared vector space. It uses raw image inputs, it is efficient and generalizable.
 %The decoder uses an auto-regressive transformer, optionally initialized from a pre-trained language model such as GPT-2 \cite{}, as a means towards improved language generalization \cite{}.
%Early image captioning approaches included predefined captions with template-based methods or retrieval-based methods.
%\subsection{Vision-Language Models}
The state-of-the-art is currently held by general purpose vision-and-language (V\&L) models \cite{li2020oscar, wang2021simvlm, hu2022scaling, li2022blip}.
%, including OSCAR \cite{li2020oscar}, SimVL \cite{wang2021simvlm}, LEMON \cite{hu2022scaling} and BLIP \cite{li2022blip}. 
These large-scale models are pre-trained on large amounts of image-text pairs to learn generic multimodal features, after which they can be finetuned to a downstream task such as image captioning, with a separately-optimized model needed for each image captioning dataset. As such, these models require extensive resources for training and deployment. %and thus have somewhat limited application.  

\begin{figure*}
    \centering
    \includegraphics[width=1\textwidth]{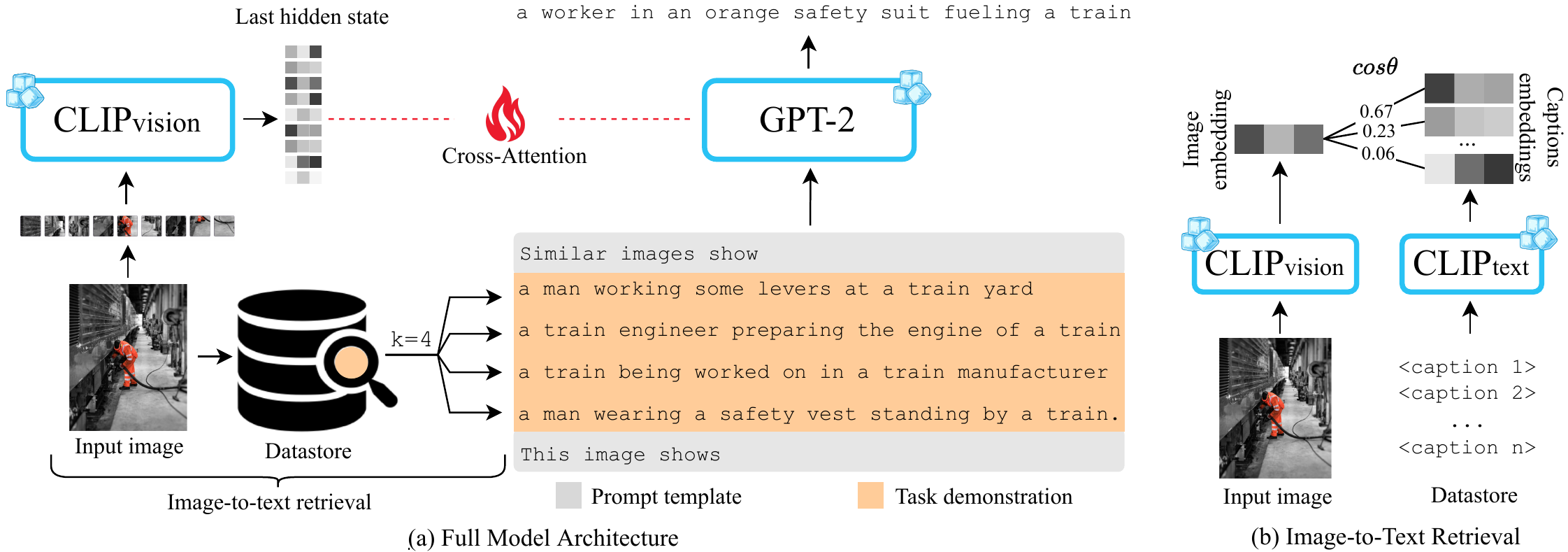}
    \caption{The \model approach to image captioning. (a) \model generates a caption conditioned on the encoded input image, as well as on a set of $k$  retrieved captions which are used as a task demonstration, input to the decoder as a prompt. (b) The $k$ captions are retrieved from a datastore of $N$ captions via image-to-text retrieval. }
    \label{fig:model}
\end{figure*}

%Given an input image, EXTRA retrieves captions from a datastore and encodes both the input image and the retrieved captions using a pretrained vision-and-language encoder. The decoder attends over both the visual and linguistic outputs, improving the quality of the generated caption.

\subsection{Freezing Image Captioning Models}

%As a means towards improved generalization, components of the image captioning model can be initialized with pre-trained weights. The entire model would then be finetuned for a given dataset. 
%A more efficient alternative is to freeze part of or all of the pre-trained parameters, adapting the model to a dataset either through prompting \cite{frozen} or by training a small amount of new parameters \cite{sung2022vladapter}. As frozen model parameters require no gradient updates, training is faster and occupies less GPU memory.

%As a means towards improved generalization, 
Components of the image captioning model can be initialized with pre-trained weights, frozen in part or completely \cite{alayrac2022flamingo}, as a way to prevent catastrophic forgetting \cite{mccloskey1989catastrophic}, i.e. to maintain good generalization. As frozen model parameters require no gradient updates, training becomes faster and occupies less GPU memory.
ClipCap and I-Tuning \cite{clipcap, luo2022tuning} are two lightweight-training image captioning models which use a pre-trained vision encoder, CLIP~\cite{radford2021learning}, and language decoder, GPT-2~\cite{radford2019language}, as frozen model components. %CLIP has two components, a vision encoder and a text encoder, trained with a contrastive vision-and-language  objective that maps images and text into a shared vector space. GPT-2 is a transformer-based auto-regressive language model. 
To map between these two independently trained components, ClipCap employs prefix-tuning, mapping a fixed-length CLIP embedding of the image into the GPT-2 language space. I-Tuning extracts \textit{visual memory embeddings} from CLIP and uses those to adjust the output hidden states of GPT-2. In \model, we also use CLIP and GPT-2, instead connected through a set of trainable
cross-attention layers. The novelty here is that \model uses retrieval augmentation to maintain performance while substantially reducing the number of trainable parameters. %because the model can generalize from in-context examples in the prompt.	

%In our work, we also use CLIP and GPT-2, instead simply connected through a small set of trainable cross-attention layers since the model does not store all the information in its parameters, by rely instead on retrieval information.

\subsection{Retrieval-Augmented Generation}

Retrieval-augmented language generation consists of conditioning generation on additional information that is retrieved from an external datastore \cite{rag}. Retrieval augmentation has been gaining traction in other tasks \cite{li2022survey,izacard2022few}, but remains largely unexplored in image captioning. Some relevant works in image captioning include \cite{app10186235, Xu2019AUG,ramos2021retrieval, sarto2022retrieval, ramos2023retrieval}.
%Zhao \etal~\cite{app10186235} who extract textual features from retrieved captions which are used for additional conditioning of the generation process. Xu \etal~\cite{Xu2019AUG} employ a generative adversarial framework guided by retrieved captions. Fei~\cite{Fei2021MemoryAugmentedIC} use retrieved words at inference time to interpolate with the decoder output distribution. Ramos \etal~\cite{ramos2021retrieval} inject information from retrieved captions into the memory state of a LSTM decoder and in the attention mechanism.
Closest to our work, Sarto \etal~\cite{sarto2022retrieval} and Ramos \etal~\cite{ramos2023retrieval}  recently proposed retrieval-augmented transformer-based captioning models that perform cross-attention over the encoded retrieved captions. Our work differs from previous work in two main ways. We employ a simple prompt-based conditioning method, wherein retrieved captions are used as a prompt to a generative language model. Moreover, we are the first to leverage retrieval augmentation for training-free domain transfer and generalization in image captioning.

\subsection{Prompting Text Generation}
Prompts have become a common way to pass additional instructions and task demonstrations to a pre-trained language model \cite{radford2019language}.
%Radford \etal~\cite{Radford2019LanguageMA} showed that a language model trained on the task of next-word-prediction can perform a range of other tasks without retraining or additional parameters, simply through a prompt like ``translate to english'', combined with a task demonstration, e.g. showing a French sentence followed by an English sentence. 
In vision-and-language learning, prompts have been used to instruct a model to perform one of multiple tasks it was trained for \cite{li2022blip}, or to apply a model to a new task in a zero-shot fashion \cite{frozen,jin2021good}. %We use prompts with input-specific task demonstrations, as a means towards retrieval augmentation in a fully supervised training setup.
We use prompts with a task demonstration tailored to the specific input image, as a means towards retrieval augmentation.

\section{Proposed Approach}
% state general model description again (from Intro)

%\textsc{SmallCap} is an image captioning model augmented with retrieval through the use of a prompt. %The model is depicted in Figure~\ref{fig:model}.

%\textsc{SmallCap} is an image captioning model augmented with retrieval through the use of a prompt. %The model is depicted in Figure~\ref{fig:model}.

\subsection{Model}

\model is a lightweight-training image captioning model augmented with retrieved captions through the use of a prompt. 
\model combines powerful pre-trained unimodal models in an encoder-decoder architecture, as shown in Figure~\ref{fig:model} (a). As encoder we use CLIP \cite{radford2021learning}, which produces a sequence of patch embeddings. As decoder we use GPT-2 \cite{radford2019language}.
%and as decoder we use GPT-2, for its strong language generation capabilities \cite{radford2019language}. 
These two models operate in different vector spaces, so we connect them with multi-head cross-attention, through which each layer of the decoder attends to the encoder outputs \cite{NIPS2017_3f5ee243}. In order to reduce the compute requirements for training and to preserve their generalization capabilities, we freeze the encoder and decoder and only train the randomly-initialized cross-attention layers between them. We %can
further control the number of trainable parameters through the dimensionality of the projection matrices in the cross-attention layers, which we denote as $d$. For GPT-2, a model  with $d_{model}=768$ hidden dimensions and $h=12$ cross-attention heads, $d$ defaults to 64 ($d_{model}/h$), as per Vaswani \etal~\cite{NIPS2017_3f5ee243}, but can be arbitrarily set to any value (see Appendix~\ref{app:cross-attention} for more details). %As shown in Vaswani \etal~\cite{NIPS2017_3f5ee243}, smaller values of $d$ can drastically improve compute efficiency while still being highly performant.

Similarly to retrieval-augmented models for other tasks~\cite{li2022survey,rag,izacard2022few, wang-etal-2022-training}, \model does not need to store all necessary information within its parameters, because it has access to external knowledge from a datastore of text. 

\subsection{Prompting with Retrieved Captions}
\label{datastore_explained}

Instead of the image-to-image retrieval methods used in recent work~\cite{sarto2022retrieval}, which are limited to image captioning data in the datastore, we employ image-to-text retrieval, as shown in Figure~\ref{fig:model}~(b). In this way, \model can make use of a datastore containing any type of text that is considered useful for describing images, be that image captions, video captions, audio captions, etc.  Here, we exploit the full CLIP model, with its vision and text encoders, which map the two modalities into a shared vector space. We encode an input image and the contents of the datastore, and use nearest neighbor search based on cosine similarity to retrieve the $k$ text items from the datastore most similar to the image. The retrieved text is used to fill the slots in a fixed prompt template of the following form: \texttt{Similar images show} \{\texttt{caption$_1$}\}\texttt{...}\{\texttt{caption$_k$\}}. \texttt{This image shows \_\_\_}.\footnote{See Appendix \ref{appendix_prompt} for more information on the prompt template.} 
The last sentence of the prompt is similar to the simple, fixed prompts used in other studies \cite{li2022blip}, but here this cue is preceded by a demonstration of the captioning task, tailored to the input image. The decoder receives this prompt as input tokens and then generates a caption conditioned on the image features \textbf{V} and the task demonstration \textbf{X}. The weights in the cross-attention layers ($\theta$) are trained by minimizing the cross-entropy loss of predicting the $M$ tokens in the reference $y_1,\dots,y_M$:
\begin{align}\label{eqatt1}
L_{\theta}=-\sum_{i=1}^{M}\log P_{\theta}(y_{i}| y_{<i}, \textbf{X}, \textbf{V}; \theta).
\end{align}
% \begin{center}
%     \begin{tabular}{l}
%       \texttt{Similar images show} \\
%       \texttt{\{caption$_1$\}}\\
%       \texttt{\{caption$_2$\}}
%       \\ \texttt{...} \\
%       \texttt{\{caption$_n$\}}\\
%       \texttt{This image shows}
%     \end{tabular}
% \end{center}

\begin{comment}
    
\begin{center}
    \begin{tabular}{l}
    
      \small\texttt{Similar images show} \\
      \small\{\texttt{caption$_1$}\}\\
      %\{\texttt{caption$_2$}\}\\
      \small... \\
      \small\{\texttt{caption$_k$\}}.\\
      \small\texttt{This image shows \_\_\_      }
    \end{tabular}
\end{center}
\end{comment}

% \begin{center}
%     \begin{tabular}{l}
      
%     \end{tabular}
% \end{center}

\begin{table}[]
    \resizebox{\linewidth}{!}{
    \centering
    \begin{tabular}{llllll}
    \toprule
        Model & $|\theta|$ & B@4 & M & CIDEr & S   \\ \midrule
        \multicolumn{6}{c}{Large Models with V\&L pre-training} \\ 
        \midrule
        
        LEMON$_{\text{Huge}}$ \cite{hu2022scaling} & 675 &  \textbf{41.5} & 30.8 & 139.1 & 24.1  \\ %Hu2021ScalingUV
        SimVLM$_{\text{Huge}}$ \cite{wang2021simvlm} & 632 & 40.6 & \textbf{33.7} & \textbf{143.3} & \textbf{25.4}  \\ % nocaps results copied from lemon paper
        OSCAR$_{\text{Large}}$ \cite{li2020oscar} & 338 & 37.4 & 30.7 & 127.8 & 23.5\\ 
        BLIP$_{{\text{CapFilt-L}}}$ \cite{li2022blip} & 224 &  39.7 &-& 133.3 &-  \\ 
        %Unified VLP & 86 & 36.5 & 28.4 & 117.7 & 21.3 & - & - & - & - \\ 
        %VL-T5 & 172 & 34.5 & 28.7 & 116.5 & 21.9 & - & - & - & - \\ %cho2021vlt5
        \midrule
        \multicolumn{6}{c}{Lightweight-training models} \\
        \midrule  
        I-Tuning$_{\text{Large}}$ \cite{luo2022tuning} & 95 & 34.8 & 29.3 & 119.4 & \textbf{22.4} \\
        CaMEL \cite{barraco2022camel}& 76 & \textbf{39.1} & \textbf{29.4} & \textbf{125.7} & 22.2  \\
        I-Tuning$_{\text{Medium}}$  \cite{luo2022tuning} & 44 & 35.5 & 28.8 &  120.0 & 22.0 \\%barraco2022camel
        ClipCap \cite{clipcap}& 43  & 33.5 & 27.5 & 113.1 & 21.1 \\
        I-Tuning$_{\text{Base}}$  \cite{luo2022tuning}& 14 & 34.8 & 28.3 & 116.7 & 21.8  \\ 
        \textsc{SmallCap}& \textbf{7}  & 37.0 & 27.9 & 119.7 & 21.3  \\
        %\midrule\midrule
        %\multicolumn{6}{c}{Model variants} \\
        \midrule  
        % \midrule
        %
        %\textsc{SmallCap} & 28  & 36.6 & 28.1 & 120.2 & 21.6  \\

        \textsc{SmallCap}$_{\text{d=16, Large}}$  & 47 & 37.2 & 28.3 &  121.8 & 21.5 \\
        %\midrule
        %\multicolumn{6}{c}{Smaller cross-attention}\\ 
        \textsc{SmallCap}$_{\text{d=16, Med}}$ & 22  & 36.5 & 28.1 & 120.7 & 21.6  \\%\midrule
        \textsc{SmallCap}$_{\text{d=8, Base}}$ & 3.6  & 36.7 & 27.8 & 119.1 & 21.1  \\
        \textsc{SmallCap}$_{\text{d=4, Base}}$ & 1.8  & 36.0 & 27.4 & 117.4 & 21.0  \\

    \bottomrule
    \end{tabular}
    }
    \caption{Results on the COCO test set with cross-entropy training. 
    $|\theta|$: number of trainable parameters in the model (in millions).% Best scores in each category are in bold, second best among lightweight models are underlined. Despite its highly reduced size, \model ranks second in its category on BLEU-4. %$^{\dagger}$only a subset of parameters trained.
    }
    \label{tab:main_results}
\end{table}

The datastore used to train \model can change from training to inference, depending on the application. For example, additional data can be added to enable better generalization, or the datastore can be entirely swapped for new data at inference time to enable domain transfer without the need for retraining, as shown in Section~\ref{sec:training}.

%To summarize, our proposed approach is light-weight training and parameter-efficient, as it is only training with multimodal data the cross-attention layers, compensated by building on top of unimodal models that leverage retrieval information. 

%In this way, the datastore %forms the core of our approach, since serves as a non-parametric external memory and enables the training-free use of data. 
%We note that when the datastore is populated solely from the data used for training \model, it serves purely as an external memory store. The data within the datastore is also \textit{seen} by the model during training, but does not need to be memorized, since that same knowledge is available through retrieval.
%Alternatively, the contents of the datastore could be swapped out for other data or enriched with additional data, enabling the training-free use of new data for the purpose of domain transfer and improved generalization. 
%Alternatively, additional data could be included in the datastore to enable better generalization, or the datastore could be entirely swapped out for new data at inference time to enable domain transfer without the need for retraining.

%\model generates a caption conditioned on the input image and text retrieved from an external datastore and presented to the decoder as a prompt. 

\section{Main Experiments}

%In this section we describe the general experimental setup, including details on the training of \model, and compare performance against state-of-the-art models.  

%We report results on the COCO image captioning benchmark \cite{} as well as on \texttt{nocaps}, a challenge dataset for evaluating the generalization capabilities of models trained on COCO \cite{}.

% section on SOTA (which is just coco)
% section on training-free domain transfer
% subsection on ID data only
% subsection on ID data + L/W/L+W
% section on domain-agnostic 

\subsection{Experimental Setup}
\label{experimental_setup}

\model's encoder and decoder are initialized respectively from CLIP-ViT-B/32 and GPT-2$_{\text{Base}}$, as available on HuggingFace \cite{wolf2020transformers}. The encoder and decoder are not updated and only the cross-attention layers between them are trained.
A 12-head cross-attention layer is added to each of the 12 layers of GPT-2.
To achieve a low number of trainable parameters, we vary the dimensionality of the projection matrices in the cross-attention layers, $d$, by scaling from the default size of 64 down to 16, 8 and 4, which results in model variants with 7M, 3.6M and 1.8M trainable parameters, respectively. Our main model, \model, has 7M trainable parameters and a total of 218M parameters (including the frozen CLIP encoder and GPT-2 decoder).

The cross-attention layers are trained on the COCO dataset  \cite{chen2015microsoft} using the standard Karpathy splits \cite{karpathy2015deep}. The models are trained to minimize the cross-entropy loss using an AdamW optimizer \cite{loshchilov2018fixing} with an initial learning rate of 1e-4 and a batch size of 64. Training runs for 10 epochs and we use the epoch checkpoint with the best CIDEr score on the validation set. Training takes up to 8 hours on a single NVIDIA A100 GPU, using 16 GB of the available memory. %During training, the model was prompted with a set of $k=4$ retrieved captions per image. 

During training, the model is prompted with a set of $k=4$ captions per image, retrieved from a datastore of the training captions from COCO. 
Retrieval is based on CLIP-ResNet-50x64\footnote{Downloaded from \url{https://github.com/openai/CLIP}} representations of input images and captions in the datastore, the latter being precomputed offline and indexed with FAISS \cite{JDH17} for efficient nearest neighbor searching.\footnote{We use an inner product index (\texttt{IndexFlatIP}) without any training and normalize the representations to search based on cosine similarity.} During inference, the model generates a caption using beam search decoding with a beam size of 3. Inference, including retrieval and prompting, takes 0.22 seconds on average across 1,000 randomly sampled images, compared to 0.19 seconds without retrieval. For more details on design choices and hyperparameters, see Appendix~\ref{system_dev}.
%, since we found it worked better for retrieval performance compared to ViT-based CLIP, as later described in section \ref{system_dev} and in line with previous work \cite{shen2021much}. 

For evaluation, we compute the standard metrics: BLEU-4 (B@4) \cite{papineni2002bleu}, METEOR (M) \cite{denkowski2014meteor}, CIDEr \cite{vedantam2015cider}, and SPICE (S) \cite{anderson2016spice}, using the COCO evaluation package.\footnote{\url{https://github.com/tylin/coco-caption}}

\subsection{Benchmark Results}
\label{results}

Here, we report results on COCO \cite{chen2015microsoft}, as well as on \texttt{nocaps}\cite{Agrawal_2019}, a challenge dataset for evaluating the generalization capabilities of models trained on COCO.

\begin{table}[]
 \centering
    \resizebox{.935\linewidth}{!}{
   
    \begin{tabular}{lllll}
    \toprule
        %& \multicolumn{4}{c}{NoCaps }  \\
        %\midrule
        Model & In & Near & Out & Entire  \\ \midrule
        OSCAR$_{\text{Large}}$$^\diamond$ &84.8&82.1&73.8& 80.9 \\ 
        CaMEL$^\star$ &  \textbf{88.1} & 79.1 & 54.6 & 75.9 \\ %barraco2022camel
        ClipCap$^\star$ & 74.5 & 65.6 & 47.1 & 63.4\\
        \textsc{SmallCap} &  83.3 & 77.1 & 65.0 & 75.8 \\ 
        \textsc{SmallCap}$_{\text{+W+H}}$ & 87.9 & \textbf{84.6} & \textbf{84.4} & \textbf{85.0}\\
    \bottomrule
    \end{tabular}
    }
    \caption{CIDEr results on the \texttt{nocaps} test set.  $\diamond$: Results copied from the respective publications. $\star$: Results computed by us. +W+H: datastore with additional Web and Human-labeled data.
    }
    \label{tab:nocaps_test}
\end{table}

\paragraph{COCO:} In Table~\ref{tab:main_results} we benchmark our approach on the COCO dataset. %We dub our main model as \model, and its smaller versions as .
%For the model's size, we explored reducing the dimensionality of the cross-attention heads ($d$), with number of parameters ranging from 7 to 1.7M, respectively.
In the top half of the table, we acknowledge the strong performance of large-scale pre-trained models, ranging in size from 224M to 675M trainable parameters. We also note that these models are pre-trained on 4M--1.8B image-caption pairs, i.e., much more than the COCO data. 

In the lower half of the table we see how our approach compares to other lightweight-training models. %Among \textsc{SmallCap}'s predecessors, I-tuning$_{medium}$ and CaMEL have the best trade-off between size and performance.
%I-tuning$_large$ has an oddly low performance considering its relatively large size, whereas I-tuning$_base$ and ClipCap lag well behind the rest. \textsc{SmallCap} performs on par with the two I-tuning$_medium$ and is outperformed only by CaMEL, a model trained end-to-end, with nearly three times as many parameters as \textsc{SmallCap}. 
With only 7M parameters, \model performs better or on par with ClipCap and I-Tuning. In this in-domain setting, it is only outperformed by CaMEL, which is trained end-to-end with eleven times as many trainable parameters.
% smaller/smallest
Reducing the number of trainable parameters to 3.6M, %\textcolor{red}{by varying the cross-attention dimension $d$}, 
\textsc{SmallCap}$_{\text{d=8, Base}}$ still yields competitive performance, and even with just 1.8M trainable parameters, \textsc{SmallCap}$_{\text{d=4, Base}}$ is better than the substantially larger models ClipCap and I-Tuning$_{\text{Base}}$. We also experiment with Medium and Large GPT-2 decoders (\textsc{SmallCap}$_{\text{Medium}}$ and \textsc{SmallCap}$_{\text{Large}}$ in Table~\ref{tab:main_results}), and find that performance scales: by one CIDEr point from Base to Medium and by another point from Medium to Large.\footnote{See Appendix~\ref{app:scaling_dec} for more results regarding scaling the decoder.} %We also show the proposed method is model agnostic by trying with a different language model, OPT \cite{zhang2022opt}, shown in Section \ref{different dec}. 
% not the best here but wait for more
Despite its small size, \textsc{SmallCap} shows competitive performance on COCO, the dataset it was trained on. %The ability of \model to leverage new data through retrieval makes it especially suitable for out-of-domain evaluation contexts, as shown in subsequent experiments.
In contrast to previous lightweight-training models, \model further has the ability to generalize and transfer out-of-domain without retraining, as shown in subsequent experiments.

\paragraph{\texttt{nocaps}:}
Results on the \texttt{nocaps} test set are reported in Table~\ref{tab:nocaps_test}.\footnote{OSCAR$_{\text{Large}}$ results with COCO-only training. CaMEL results with CLIP-ResNet-50×16, $\lambda_kd$ = 0.1, no mesh connectivity, and a cross-entropy objective (checkpoint obtained through personal communication).}$^{,}$\footnote{We only include results from models which follow the \texttt{nocaps} guidelines to not train on image-caption pairs beyond COCO \cite{Agrawal_2019}. As we use only captions for retrieval, our method is also in line with these guidelines. }$^{,}$\footnote{We also include results on the validation set in Appendix \ref{nocaps:val}.}. %In spite of being the smallest model, 
\model clearly outperforms other lightweight methods \textit{Out}-of-domain and achieves competitive performance \textit{In}-domain and \textit{Near}-domain. The model's strong generalization capabilities point to it being less prone to over-fitting as it does not need to memorize its training data, available also through retrieval. 
%Notice that OSCAR$_{Large}$, the only model that outperforms \model, benefits from pre-training on diverse image-text pairs and from using constrained beam search.
Our model can further improve when additional data is placed in the datastore, as seen in \textsc{SmallCap}$_{\text{+W+H}}$. In this variant, described in more detail in Section~\ref{subsec:largedata}, the COCO datastore is augmented with diverse web (W) and human-labeled (H) data. \textsc{SmallCap}$_{\text{+W+H}}$ shows impressive generalization capabilities, outperforming the much larger OSCAR$_{\text{Large}}$ by over 10 points in the \textit{Out}-of-domain setting.  %The strong performance of our model on \texttt{nocaps} demonstrates that it is capable of generalization to unseen visual concepts. 
Next, we further explore \model's ability for training-free transfer to new domains on diverse datasets.

%---------------------------------------------------------------------------

\section{Training-Free Use of Data}
\label{sec:training}
%[This section refers the reader to the top half of table 3, i.e. Exploration]
In this section, we study \model's ability to leverage new data in its datastore in a training-free manner, i.e. all experiments presented here constitute changes made to the datastore at inference time, while the model, trained on COCO, remains fixed. The focus is on out-of-domain performance as measured on a diverse set of captioning datasets: Flick30k \cite{young2014image}, VizWiz \cite{gurari2020captioning} and MSR-VTT \cite{xu2016msr}. The latter is in fact a video captioning dataset, which we adapt by converting video clips into an image of four 4 frames, sampled at 0, 25, 50 and 100\% of the clip duration (see the MSR-VTT example in Figure~\ref{fig:domain}). %Here, we explore the effect of using different types of data to populate the datastore.  %, and then report comparisons to state-of-the-art with the best datastore configuration found during the exploration.
We start by exploring different configurations of the datastore, with the results in Table~\ref{tab:training_free} reported on validation data.

\subsection{In-domain Data}

In the top of Table~\ref{tab:training_free}, we show how \textsc{SmallCap} performs when its datastore is populated with the training data associated with each respective dataset (\textit{In-domain}). In comparison to using COCO captions in the datastore (\textit{COCO}), the model performance substantially increases for all three datasets. This shows that \textsc{SmallCap} adapts to the retrieved information to achieve domain transfer. The improvement is most notable for VizWiz, likely because the nature of this dataset is very distinct from COCO, and thus there is a larger domain gap to be closed. %The performance on these three benchmarks can further improve when the datastore is augmented with diverse web and human-labeled data, as shown in the next experiment.

\begin{table}[]
    \centering
    \resizebox{.9\linewidth}{!}{
    \begin{tabular}{llll}
        \toprule
         \model datastore & F30K & VW & MV \\ 
        \midrule
        COCO & 52.2 & 34.5  & 23.3  \\ 
        %\midrule
        %\midrule
        In-domain & \underline{55.4} & \underline{47.7} & \underline{29.2} \\ 
          \midrule
         \multicolumn{4}{c}{Datastore augmentation}\\
           \midrule
        %In-domain + Web Raw \\
        In-domain + \textbf{W}eb & \textbf{58.6}& \textbf{48.0} & 29.8\\
        %~~~~~~\small (CC+C12M+SBU $\sim$ 12M) \\
        In-domain + \textbf{H}uman-labeled & 57.6 & 47.5 & \textbf{30.9}\\
        %~~~~~~\small (Captions, Videos, Audios, Narratives $\sim$ 2M) \\
        In-domain + \textbf{W} + \textbf{H} & 57.9 & \textbf{48.0} & 30.7\\
        \midrule
        \multicolumn{4}{c}{Domain-agnostic}\\
        \midrule

        %Web Raw  & 53.6 & 36.5 & 25.0  \\  
        %~~~~~~\small (CC+C12M+SBU) \\
        Web  & \underline{58.4}& \underline{42.4} &27.6\\
        % ~~~~~~\small (CC+C12M+SBU)  \\
        Human-labeled & 56.6 & 36.4 & 29.0 \\
        %~~~~~~\small (excl. in-domain data) \\
        Web + Human-labeled   & 57.8 & 42.2& \underline{29.9}\\
        
        %\circled{3} + \circled{1}  \\

        \bottomrule
    \end{tabular}}
    \caption{Exploration of the training-free use of data. Validation performance of \textsc{SmallCap} measured in CIDEr score, with different contents of the datastore, without any finetuning on Flickr30k (F30K), VizWiz (VW), and MSR-VTT (MV). The best number per section is underlined; the best number overall is in bold. } 
    \label{tab:training_free}
\end{table}

\subsection{Augmenting the Datastore}
\label{subsec:largedata}

In Table~\ref{tab:training_free} (\emph{Datastore augmentation}), we augment the in-domain datastore with additional large-scale data in an effort to improve generalization. %test \textsc{SmallCap}'s ability to leverage large data in a training-free fashion. %\footnote{We measured performance on COCO and found that this dataset does not benefit from any datastore augmentations so we omit those results.} 
We experiment with diverse web data (which is large-scale but automatically labeled) and human-labeled data (smaller-scale but clean).\footnote{Data size and further details can be found in Appendix~\ref{appendix_data}.}  %, with the goal of determining which type of data is more effective for augmenting the datastore.

\paragraph{+ Web Data:} We first consider large-scale data from the web, expanding the datastore with text from three web datasets \cite{li2022blip} (Conceptual Captions \cite{sharma2018conceptual}, Conceptual 12M \cite{changpinyo2021conceptual}, and SBU captions \cite{ordonez2011im2text}).\footnote{We use a trained FAISS index (\texttt{IndexIVFFlat}) for faster search.} The results with \textit{In-domain + Web} in Table~\ref{tab:training_free} show that performance improves for all three datasets. We can see a bigger improvement on Flickr30K and MSR-VTT when using a large and diverse datastore compared to just using in-domain data. Improvement on VizWiz, on the other hand, remains low, in line with the earlier observation that this dataset has a distinct distribution that is not easily matched by other data.  

\paragraph{+ Human-labeled Data:} We also consider smaller-scale but clean human-labeled data. As discussed in Section~\ref{datastore_explained}, the datastore can contain any type of text that can be useful to describe images, thus not being constrained by the assumption of image-caption pairs. As such, we consider text not only from image captions (COCO \cite{chen2015microsoft}, Flickr30k \cite{young2014image}, VizWiz \cite{gurari2020captioning}), but also from video captions (MSR-VTT \cite{xu2016msr}, VATEX \cite{wang2019vatex}, TGIF \cite{li2016tgif}), audio captions (Clotho \cite{drossos2020clotho}), and localized narratives (LN ADE20k, LN COCO, LN Flickr30k, LN OpenImages \cite{pont2020connecting}). 

As seen in \textit{In-domain + Human-labeled}, adding human-labeled data to the datastore leads to an improvement over using in-domain data only for Flickr30k and MSR-VTT but not for VizWiz. In comparison to \textit{In-domain + Web}, this improvement is smaller for Flickr30k, but larger for MSR-VTT. 
%This makes sense since the web data used in these experiments corresponds to large image captioning datasets (Conceptual Captions and SBU), thus containing more language-diverse and geographically-diverse concepts. 
Although smaller than web data, human-labeled data benefits MSR-VTT more, because it contains text from different tasks, including video captioning. 

\paragraph{+ Web + Human-labeled Data:}

Seeing that \model can benefit both from Web and from Human-labeled data as augmentations over in-domain data alone, we also consider a combination of the two, to determine whether their contributions are complementary or overlapping. The results for \textit{In-domain + W + H} in Table~\ref{tab:training_free} show that combining the two sources of data is not beneficial for any of the three datasets.

%The experiments discussed above confirm the effectiveness of augmenting an in-domain datastore with additional data for training-free domain transfer. Yet, in real-world applications, in-domain data may not always be available.

\subsection{Domain-agnostic Datastore}
\label{varying_datastore}
%These experiments have showed \model can indeed adopt to other domains without finetuning by simply replacing the datastore with captions from the target-domain and by leveraging more data. 

In this section, we study whether \model could still perform well without access to in-domain data and report results under the heading \emph{Domain-agnostic} in Table~\ref{tab:training_free}. We find that the patterns observed above with in-domain data largely hold without it as well. With the large and diverse \emph{Web} datastore, \textsc{SmallCap} performs close to or even better than with \textit{In-domain} data.  \emph{Human-labeled} data is again seen to benefit MSR-VTT the most, the optimal configuration for this dataset being \emph{Web + Human-labeled}.
%Results on MSR-VTT indicate that \emph{Web + Human-Labeled} performs better than only \emph{Web}, in line with the earlier observations. %showing that adding more data that captioning data can be beneficial for other domains like video captioning. 

From the exploration presented above, we conclude that \model's image captioning capabilities can transfer with access to web data in addition to or in place of in-domain data. The model can also leverage human-labeled data beyond image-captioning pairs in solving tasks other than image captioning, such as video captioning. 

\begin{table}[]
\centering
    \resizebox{.95\linewidth}{!}{
    
    \begin{tabular}{llll}
        \toprule
         & Flickr30K & VizWiz & MSR-VTT \\ \midrule
        ClipCap & 41.2 & 28.3 &12.5 \\
        CaMEL & 55.2 & 37.6 &  20.7 \\
        \model & \textbf{60.6} & \textbf{55.0} & \textbf{28.4}\\
        \midrule
        \multicolumn{4}{c}{Pre-training \& finetuning} \\
        \midrule
        \textcolor{gray}{SOTA} & \textcolor{gray}{79.6} \cite{luo2022vc} & \textcolor{gray}{120.8} \cite{wang2022git} & \textcolor{gray}{75.9} \cite{wang2022git} \\
        \bottomrule
    \end{tabular}
    }
    \caption{Out-of-domain performance without additional training, measured in CIDEr score on the test data. Flickr30K and VizWiz results with \emph{In-domain + Web}, and MSR-VTT result with \emph{In-domain + Human-labeled}.  We include SOTA results from large-scale pre-trained models, finetuned on the respective datasets.} 
    \label{tab:other_models}
\end{table}

\begin{figure}
    \centering
    \includegraphics[width=\linewidth]{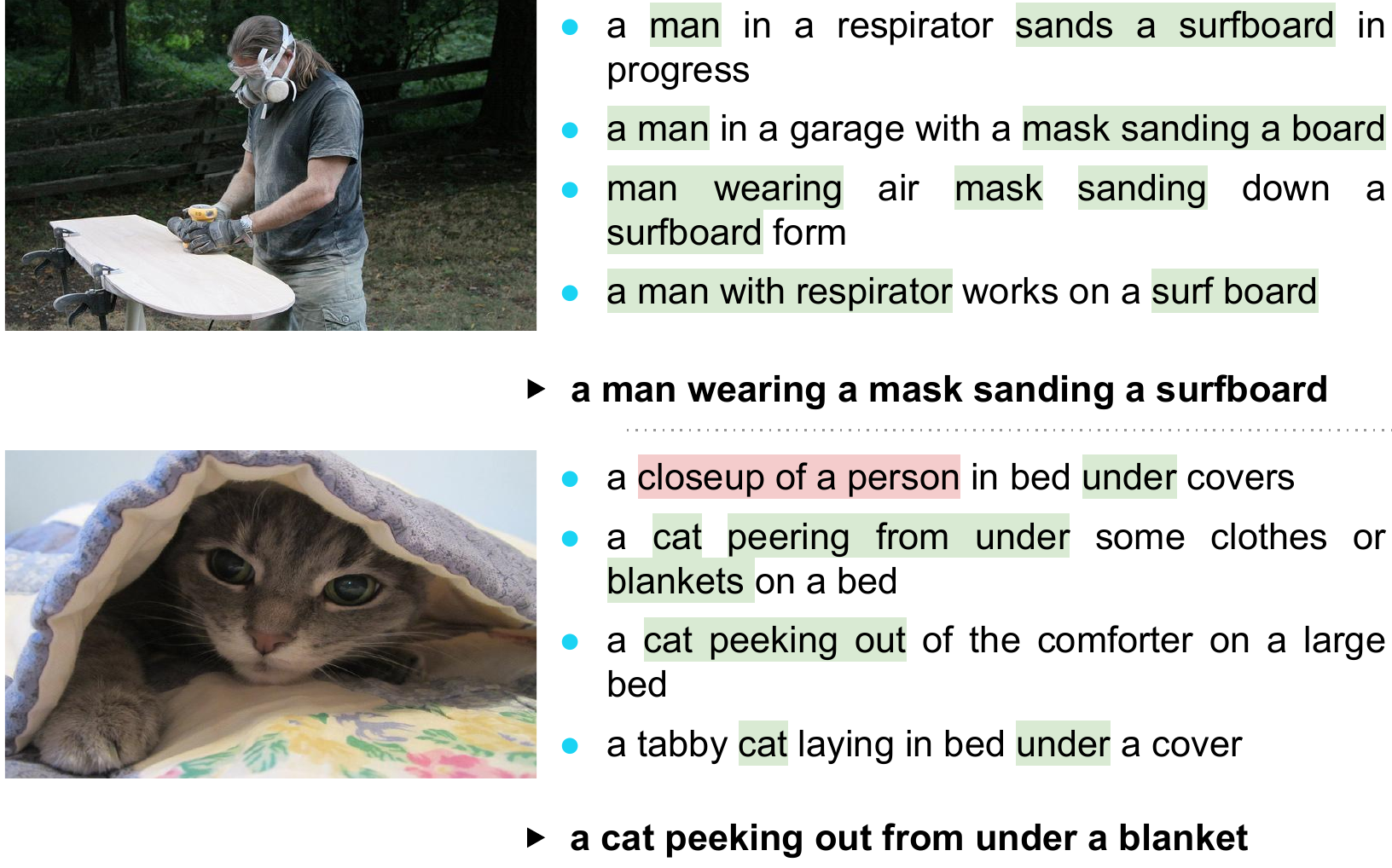}
    \caption{Examples generated by \model, together with the retrieved predictions from the COCO datastore. \cococolor{$\bullet$} denotes the retrieved captions, highlighted as green or red to indicate correct and mismatch captions, respectively. $\blacktriangleright$ denotes the generated caption.}
    \label{fig:coco_preds}
\end{figure}

\subsection{Results with the Best Configuration}

Having explored different datastore configurations for each of the three datasets, we use the best configuration for each to compare zero-shot performance against ClipCap and CaMEL, both models also trained only on COCO. In Table~\ref{tab:other_models} we show test set performance (in CIDEr score) with a datastore consisting of \emph{In-domain + Web} for Flickr30k and VizWiz, and \emph{In-domain + Human-labeled} for MSR-VTT. \model outperforms both ClipCap and CaMEL by a large margin on all three datasets. In comparison to CaMEL, the stronger baseline of the two, we see a 5.4 point improvement on Flickr30k, a noteworthy 17.4 point improvement on VizWiz and an increase of 7.7 points on MSR-VTT. The large improvement on VizWiz demonstrates \model's ability to transfer to domains very distinct from the training data, i.e., COCO. The improvement on MSR-VTT, on the other hand, shows our approach has potential not only for other domains but for other tasks as well. These results show that while other lightweight-training models lack out-of-domain generalization without finetuning, our model can transfer across domains by only swapping the datastore contents. In the bottom of the table, we provide state-of-the-art results for context, which were achieved by large-scale pre-trained V\&L models, finetuned specifically on the respective datasets.   

\section{Discussion}
%We further analyse and discuss how the retrieved information contributes to the performance of \model. 

\subsection{Qualitative Examples}

\begin{figure*}
    \centering
    \includegraphics[width=0.99\textwidth]{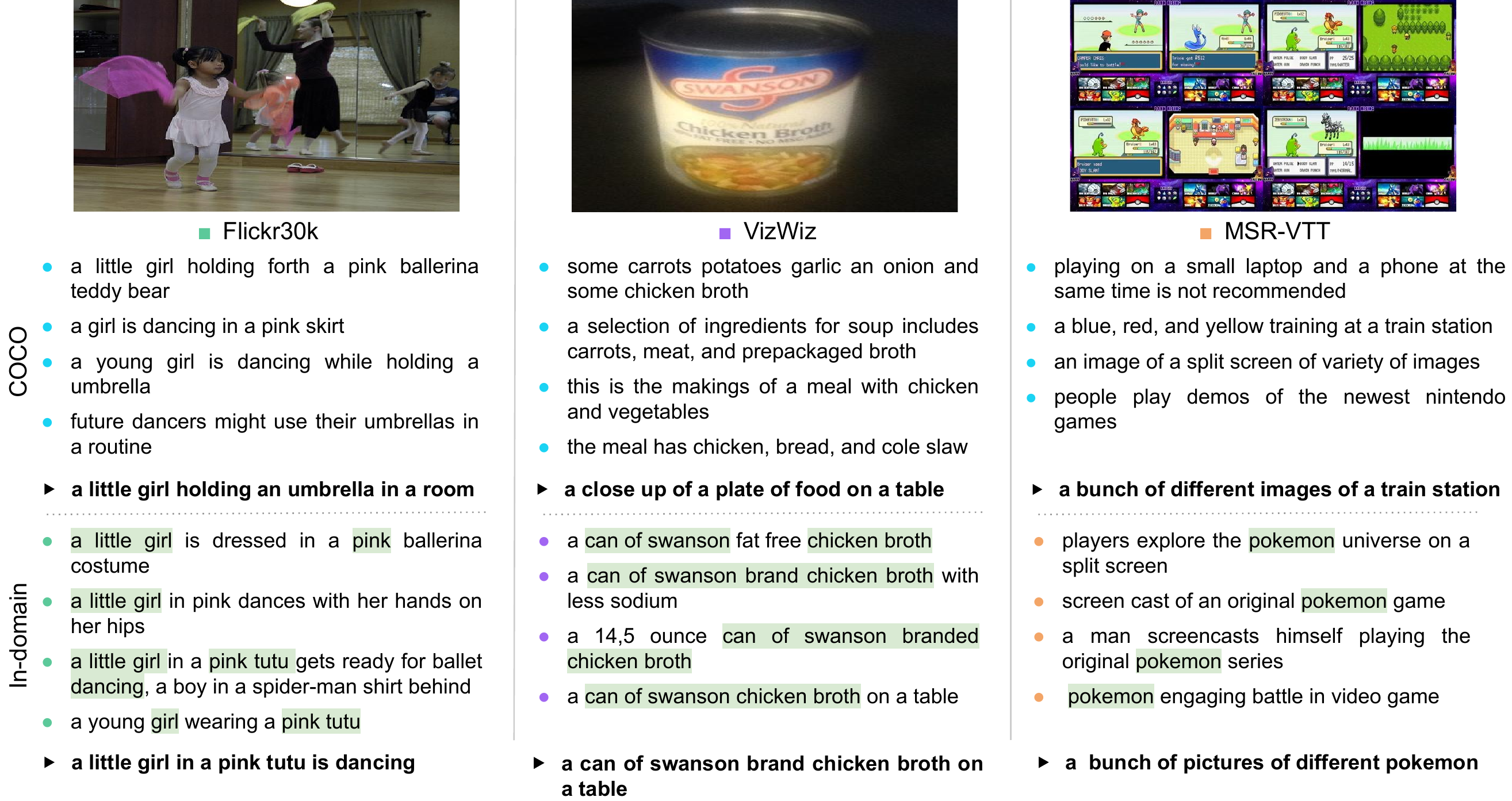}
    \caption{Examples of captions generated for Flickr30k, VizWiz and MSR-VTT, with retrieval either from COCO or in-domain data. The captions use words retrieved from the in-domain datastores which were rarely seen in the COCO training data (tutu, swanson, pokemon).}
    \label{fig:domain}
\end{figure*}

Figure~\ref{fig:coco_preds} shows examples of the retrieved and generated captions for two images from COCO. In the first example, we observe that the retrieved captions are highly relevant to the input image and the generated captions are semantically similar to them. As seen in the second example, \textsc{SmallCap} can also be robust to misleading information from retrieval. % in the presence of mismatch from certain captions.
Figure~\ref{fig:domain} shows examples of captions generated for Flickr30k, VizWiz, and MSR-VTT, with a datastore populated with COCO or with in-domain data. These qualitative results show how \model adapts to new domains: %for the Flickr30k image \model is less biased towards the very frequent concept \textit{horse} compared to the correct concept, \textit{camel}; for the VizWiz image it is more descriptive and uses the Swanson brand name,  never observed in the COCO training data, and for the MSR-VTT image it refers to Pokemon, a concept observed just six times in the COCO training data.%\textcolor{red}{for the Flickr30k image \model generates \textit{tutu}, a concept never observed in the COCO training data; for the VizWiz image it is more descriptive and uses the Swanson brand name,  also not present in COCO, and for the MSR-VTT image it refers to Pokemon, a concept observed just six times in the COCO}.
with the help of the retrieved captions, it correctly refers to the  concepts \textit{tutu}, the \textit{Swanson} brand name, and \textit{Pokemon}. The first two concepts are not present in the COCO training data at all, while the last is seen just six times.

%by simply replacing the datastore without retraining. 

%and describe concepts that were rarely seen in COCO by considering the target-domain information con.

%can adapt to new domains and can 

%can generalize 

%

%For instance, we can see 

\subsection{Analysis of the Retrieved Captions}

\begin{figure}
  \centering
  
  \resizebox{\linewidth}{!}{
  
  \begin{tabular}{@{}c@{}}
  \vspace{-6mm}
    \includegraphics[]{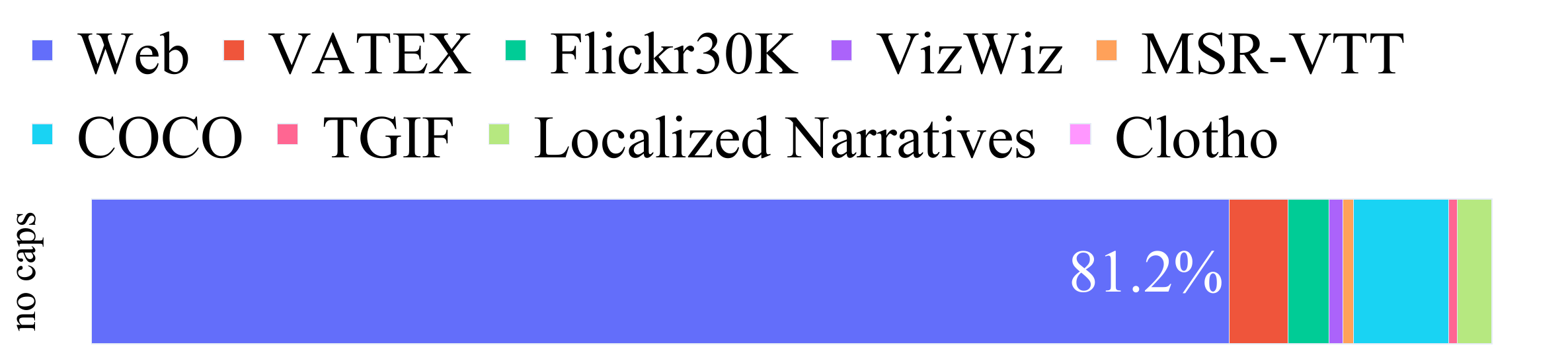} \\[\abovecaptionskip]
  \end{tabular}
  }
  \vspace{-2mm}
\resizebox{\linewidth}{!}{
  \begin{tabular}{@{}c@{}}
    \includegraphics[]{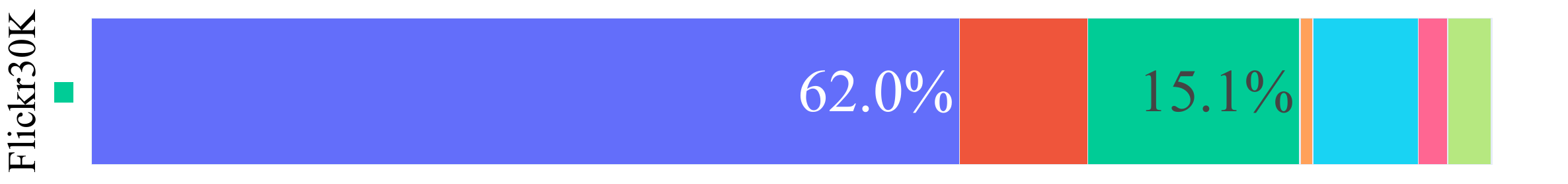} \\[\abovecaptionskip]
  \end{tabular}
  }
  \vspace{-2mm}
  \resizebox{\linewidth}{!}{
  \begin{tabular}{@{}c@{}}
    \includegraphics[]{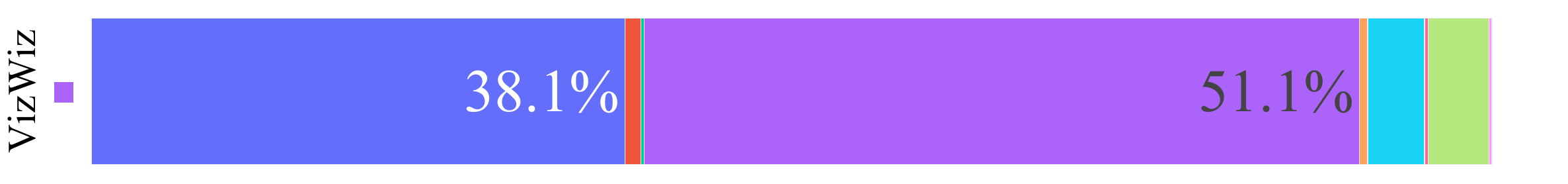} \\[\abovecaptionskip]
  \end{tabular}
  }
  %\vspace{-2mm}
  \resizebox{\linewidth}{!}{
  \begin{tabular}{@{}c@{}}
    \includegraphics[]{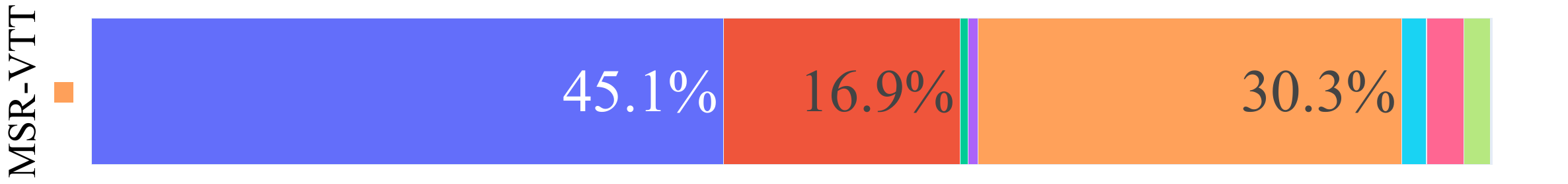} \\[\abovecaptionskip]
  \end{tabular}
  }
  \caption{Percentage of the retrieved captions that come from each data source, when testing the model on the different benchmarks of \texttt{nocaps} , Flickr30k, VizWiz and MSR-VTT.}\label{fig:percentage}
  \label{fig:stackbar}
\end{figure}

In Section~\ref{varying_datastore}, we demonstrated the ability of \model to exploit large data in a training-free fashion. Here, we inspect the distribution of retrieved captions in the \textit{In-domain + Web + Human-labeled} setting, in order to understand the individual impact of each dataset. As can be seen in Figure~\ref{fig:percentage}, most text is retrieved from web data, especially in the presence of unseen visual concepts, as is the case for \texttt{nocaps}. 
Besides web data, the model tends to retrieve text from the corresponding dataset or from a similar domain; for instance, MSR-VTT retrieval also relies on other video datasets. Due to its unique distribution, VizWiz stands out as the case with the highest rate of in-domain retrieval.

Seeing that text from all types of human-labeled data is retrieved, we measure the actual impact of each type on performance. In Table~\ref{tab:human_labeled_data}, we report performance on Flickr30k, VizWiz, and MSR-VTT, with an in-domain datastore augmented with either Image captions, Video captions, localized Narratives, or Audio captions. We see
that \model can indeed benefit from data beyond image captions. 
For instance, video captions help not only for MSR-VTT, but also for Flickr30k and VizWiz. Flickr30k benefits the most from localized narratives since this dataset contains narratives for the Flickr30k images. 
Audio captions are beneficial for both Flickr30k and MSR-VTT. Considering the distinct nature of the audio and visual modalities, this finding demonstrates the potential of leveraging data which has previously seen limited application to image captioning.

\begin{table}[]
\centering
\resizebox{.92\linewidth}{!}{
    
    \begin{tabular}{lccc}
        \toprule
         & Flickr30K & VizWiz & MSR-VTT \\ \midrule
        In-domain & 52.2 & 47.7 & 29.2 \\
        ~ + Image & 56.7 & 47.8&  29.8 \\
        %~~~~~~\small (COCO, F30K, VW)  \\
        ~ + Video & 57.0 & 47.8 &  31.1 \\
        %~~~\small (COCO, F30K, VW)  \\
        ~ + Narratives & 57.1 & 47.2&  28.7 \\
       % ~~~\small (COCO, F30K, VW) \\
        ~ + Audio & 55.4 & 47.7 &  29.4 \\
       % ~~~\small (COCO, F30K, VW) \\
        \bottomrule
    \end{tabular}
    }
    \caption{\model performance with retrieval from the different sources of the Human-labeled data. The model can benefit from having access to text that is not only from image captioning tasks, but also from other tasks such as audio captioning.} 
    \label{tab:human_labeled_data}
\end{table}

%We were what is the impact \see section b. 

%\subsection{Analysis of the retrieved performance}
%We have in seen in the previous section that the retrieved text can come more from one dataset than another, such as Web Synthetic. However, it is important to note that the dataset where the captions are retrieved from does not mean it is the actually the best dataset to be retrieved from, since the retrieval system is prone to error. We thus study the actual impact on performance of Flick30k validaton set of each the different datasets introduced in \secion \ref{varying_datastore}, reported in Table \ref{}. We can observe that \smallcap can indeed benefit from including in the datastore text beyond image-captions pairs, such as TGIF, video captioning and audio captioning, that has of before seen limited applications to image captioning.

%https://docs.google.com/drawings/d/1KQ55x9mIMDy8oLB4p1kuRSigyeyMsJl4LnIRrtrIxoE/edit

% on coco: show some retrieved captions/good case/bad case

% on Flickr: we show images with retrieved captions and predictions for COCO DS, and in-domain
\begin{figure}[]
    \centering
    \includegraphics[width=1.0\linewidth]{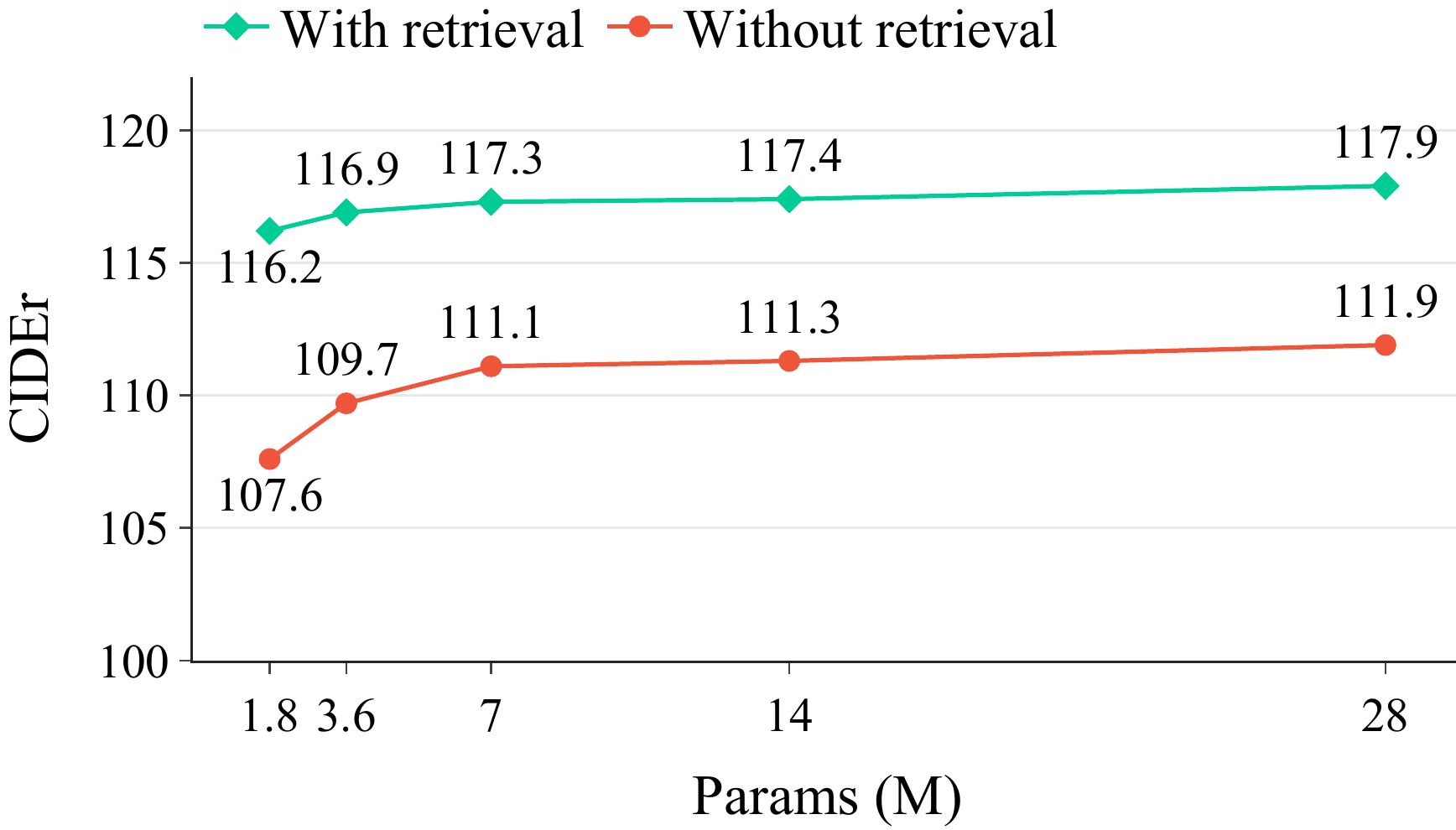}
    \caption{CIDEr scores on the COCO validation set, with and without retrieval, across different cross-attention sizes.}
    \label{fig:decay_cross_attention}
\end{figure}

\subsection{The Impact of Retrieval}
\label{cross_attention_reduction}
%Finding that the aforementioned hyperparameters for retrieving gave our model a ceiling performance of 117.7 CIDEr by training only 28M parameters, we were then interested if the number of trainable parameters could be further reduced while keeping a similar performance. 
%In this subsection, we demonstrate that the success of \model stems largely from retrieval augmentation, and that the optimal model size is 7M parameters.
In Figure~\ref{fig:decay_cross_attention}, we show validation performance with 1.8, 3.6, 7, 14 and 28 million trainable parameters with and without retrieval augmentation.\footnote{The model sizes correspond to $d=4, 8, 16, 32$ and $64$.} For variants with retrieval augmentation, performance is stable across the range of model sizes considered. Reducing the number of trainable parameters by a factor of four, from 28M to 7M, leads to a slight drop of 0.6 CIDEr points. This indicates that \model has a close-to-optimal size to performance trade-off. 

Next, we ablate the retrieval augmentation to quantify its impact. We train models without retrieval augmentation, prompting them with just the phrase \texttt{This image shows}. %instead of the template presented in Section~\ref{datastore_explained}. 
As seen in Figure~\ref{fig:decay_cross_attention}, without the aid of retrieved captions, there is a notable drop in performance compared to results with retrieval. Moreover, model performance degrades at a higher rate: while performance at the two extremes of model sizes differs by just 1.7 CIDEr points with retrieval, without it the difference is 4.3 points.%This confirms the key role of retrieval augmentation in compensating for the reduced model size by giving \model access to non-parametric knowledge.
\footnote{See Appendix~\ref{examplas_without_retrieval} for qualitative examples with and without retrieval.}

\begin{table}[]
\centering
    \resizebox{\linewidth}{!}{
    
    \begin{tabular}{llllll}
    \toprule
        Decoder & $|\theta|$ & B@4 & M & CIDEr & S   \\ \midrule
        
       % \textsc{GPT-2}& 7  & 37.0 & 27.9 & 119.7 & 21.3  \\
       
        %\midrule  
           GPT2-Base$_{\text{d=16}}$ & 7  & 37.0 & 27.9 & 119.7 & 21.3  \\
           OPT-125M$_{\text{d=16}}$ &7  & 37.6 & 28.4 &  122.0 & 21.7 \\
         \midrule
        GPT2-Medium$_{\text{d=16}}$  &22  & 36.5 & 28.1 & 120.7 & 21.6  \\      
         OPT-350M$_{\text{d=16}}$  &22 & 37.5 & 28.7 &  122.7 & 22.0 \\
    \bottomrule
    \end{tabular}
    }
    \caption{Results with different decoders on the COCO test set.% Best scores in each category are in bold, second best among lightweight models are underlined. Despite its highly reduced size, \model ranks second in its category on BLEU-4. %$^{\dagger}$only a subset of parameters trained.
    }
    \label{tab:diff_dec}
\end{table}

%Observing the importance of retrieval, 
In order to confirm that \model is not simply paraphrasing the retrieved captions without attending to the visual input, we experiment with ablating the visual modality. For this, we train a model on ``blank'' input images, setting the visual features from the encoder to zero. This yields a much lower CIDEr score of 90.1 on the validation set, showing that \model indeed uses the visual input.

\subsection{Alternative Decoders}
\label{different dec}

At the request of the anonymous reviewers, we include additional experiments with a more recent language model. Here, we use OPT-125M and 350M \cite{zhang2022opt}, equivalent in size to GPT2-Base and Medium.\footnote{There is no OPT variant equivalent in size to GPT2-Large.} The results in Table~\ref{tab:diff_dec} show that \model also performs well with these language models and is therefore model agnostic.\footnote{See Appendix~\ref{app:scaling_dec} for OPT results without retrieval.}$^{,}$\footnote{Due to our academic computing budget, we only repeat the experiments from Table~\ref{tab:main_results}. Future work can experiment further in this direction.}% for these language models as well. %Future work should investigate the potential for more recent language models in this framework.  

\section{Conclusion}

%In this paper, we explore retrieval augmention as a mean to accomplish a lightweight-training captioning model that can perform on a variate of domains without the need for retraining. Our model, SmallCap, 

In this paper, we propose \model, an image captioning model augmented with retrieval, which is light to train and can be transferred across domains without retraining. %Having access to retrieved information, %our model requires a reduced number of trainable parameters %and can transfer across domains without finetuning.
%\model demonstrates (i) lightweight training, (ii) training-free domain transfer, and (iii) exploitation of large data in a training-free fashion. 
Results on the COCO dataset show that \model is competitive to other lightweight-training models despite having substantially less trainable parameters, instead leveraging non-parametric information from a datastore of text. Out-of-domain evaluations show that \model can also perform training-free domain transfer when given access to a datastore with target-domain data. Our model further benefits from diverse web and human-labeled data in addition to or in place of target-domain data. We find that \model benefits not just from access to image captions, but also to video and audio captions (resources neglected in image captioning work in the past). 

\model's small size and impressive performance in out-of-domain settings attest to the potential of retrieval augmentation as an alternative to the expensive training found in large pre-trained vision-and-language models and the costly finetuning that even previous lightweight-training models require in order to adapt to different image captioning datasets. Future work can apply our retrieval augmentation approach to a wider range of multimodal tasks, and further explore the scalability of the data used for retrieval.

\section*{Acknowledgements}
This research was supported by the Portuguese Recovery and Resilience Plan through project C645008882-00000055, through Funda\c{c}\~ao para a Ci\^encia e Tecnologia (FCT) with the Ph.D. scholarship 2020.06106.BD, and through the INESC-ID multi-annual funding from the PIDDAC programme (UIDB/50021/2020). 


{\small
\bibliographystyle{ieee_fullname}
\bibliography{egbib, anthology,custom}
}

\appendix

\newpage

\section{Cross-Attention Layers}
\label{app:cross-attention}
We add cross-attention layers to GPT-2 following Vaswani \etal~\cite{NIPS2017_3f5ee243}. A set of queries $Q$, values $V$, and keys $K$ are processed by multi-head cross-attention (MHA) with $h$ heads as follows:

\begin{equation}
    %\begin{align}
    \text{MHA}(\textbf{Q}, \textbf{K}, \textbf{V} ) =
    \text{Concat}(\text{head}_i, ..., \text{head}_h)\textbf{W}_O,
    %\end{align}
\end{equation}

\begin{equation}
\text{head}_i=\text{Att}(\textbf{QW}_i^Q,\textbf{KW}_i^K,\textbf{VW}_i^V),
\end{equation}

 \begin{equation}
 \text{Att}(\textbf{Q},\textbf{K},\textbf{V})= \text{softmax}\frac{\textbf{QK}^T}{\sqrt{d_k}}\textbf{V},
 \end{equation}

\noindent where $\textbf{W}_i^K \in \mathbb{R}^{d_{encoder} \times d}$, $\textbf{W}_i^V \in \mathbb{R}^{d_{encoder} \times d}$,  $\textbf{W}_i^Q \in \mathbb{R}^{d_{decoder} \times d}$, and $\textbf{W}_O \in \mathbb{R}^{h*d \times d_{decoder}}$ are learned model parameters, and the attention dimensionality $d$ is set manually to a desired value.

We explore different values for the dimensionality of the cross-attention projection matrices ($d$) to achieve a lower number of trainable parameters, as discussed in Section~\ref{cross_attention_reduction}.

\section{Design Choices and Hyperparameters}
\label{system_dev}

\begin{table}[!h]
  \centering
  \subfloat[CLIP version for retrieval]{%
    \hspace{.2cm}%
    \begin{tabular}{ll}
        \toprule
        Retrieval encoder & CIDEr \\
        \midrule
        ViT-B/32 & 109.5\\
        ViT-L/14 & 115.2 \\
        ResNet-50x4 & 114.1\\
        ResNet-50x64 & 117.9 \\    
        \bottomrule
    \end{tabular}%
    \hspace{.2cm}%
  }\hspace{.2cm}
  \subfloat[CLIP version in main model]{%
    \hspace{.2cm}%
    \begin{tabular}{ll}
        \toprule
        Image encoder & CIDEr \\
        \midrule
        ViT-B/32  & 117.9 \\
        ResNet-50x64 & 107.5 \\    
        \bottomrule
    \end{tabular}%
    \hspace{.2cm}%
  }\hspace{.2cm}
  \subfloat[k for retrieval]{%
    \hspace{.2cm}%
    \begin{tabular}{ll}
        \toprule
        k & CIDEr  \\
        \midrule
        1 & 113.38\\
        2 & 116.03\\
        3 & 117.47\\
        4 & 117.88\\
        5 & 117.87\\
        6 & 117.82\\
        \bottomrule
    \end{tabular}%
    \hspace{.2cm}%
  }
  \caption{Hyperparameter tuning of the retrieval mechanism, measuring CIDEr on the validation set of COCO.}
  \label{tab:rsb}
\end{table}

%This section describes how we arrived at the configuration of \model presented in Section~\ref{experimental_setup}, as tuned on the COCO validation set. 
We developed the optimal configuration for \model by first tuning the retrieval encoder, then the main vision encoder, followed by the number of retrieved captions, and lastly the cross-attention dimensionality. The results from the first three steps are presented below, while the last step is presented in Section~\ref{cross_attention_reduction}. At the start of the tuning process, the main vision encoder was set to CLIP-ViT-B/32, the number of retrieved captions to 5 and the cross-attention dimensionality to 64.

\subsection{Retrieval Encoder}
We  compared three CLIP versions for retrieval. As seen in Table~\ref{tab:rsb} (a), CLIP-ResNet-50x64 performs best so we used this encoder for the final \model model.  

\subsection{Main Vision Encoder}
Next, we compared the use of CLIP-ResNet-50x64 to CLIP-ViT-B/32 as vision encoder in the main model. To use CLIP-ResNet-50x64 as an image encoder, we added a linear projection to match the dimensionality of the encoder to that of the decoder for the purposes of cross-attention. In Table~\ref{tab:rsb} (b), we see that CLIP-ViT-B/32 has better performance. %This linear projection was not trained to avoid adding an increase number of model parameters, which can explain why CLIP-ResNet-50x64 had a poor performance as an image encoder.

\subsection{Number of Retrieved Captions}
We also tuned the number of retrieved captions, training the model with $k$ ranging from 1 to 6. Results are reported in Table~\ref{tab:rsb} (c) and indicate that $k=4$ is the optimal value. Qualitative analysis also showed that it is important to retrieve a sufficient number of captions since %retrieving fewer captions can hurt captioning performance, whereas 
retrieving more captions can make the model more robust against wrong information from certain retrieved captions, as depicted in the second example in Figure~\ref{fig:coco_preds}.

\section{Prompt}
\label{appendix_prompt}
Besides the template proposed in Section~\ref{datastore_explained}, we explored other templates for prompting, including different separators between the retrieved captions (e.g., comma, dot, empty lines). However, we found that the prompt template has little impact on the model's performance, in line with previous work \cite{jin2021good}. Our final template was:  
\newline

\noindent\texttt{Similar images show\textbackslash n\textbackslash n\texttt{<caption 1>}\textbackslash n\textbackslash n<caption 2>\textbackslash n\textbackslash n\texttt{<caption 3>}\textbackslash n\textbackslash n\texttt{<caption 4>}.\textbackslash n\textbackslash nThis image shows}

\section{\texttt{nocaps}}
\label{nocaps:val}

\begin{table}[!h]
     \resizebox{.935\linewidth}{!}{
     \centering
     \begin{tabular}{lllll}
     \toprule
         %& \multicolumn{4}{c}{NoCaps }  \\
         %\midrule
         Models & In & Near & Out & Entire  \\ \midrule
         \multicolumn{5}{c}{Validation} \\ 
         \midrule
         OSCAR$_{\text{Large}}$* & 78.8 & 78.9 & 77.4 & 78.6\\ 
         I-Tuning$_{\text{Large}}$$^\circ$ & 89.6 & 80.4 & 64.8 & 78.5\\
         %CAMeL$^\star$ &   \\
         I-Tuning$_{\text{Medium}}$$^\circ$ & 89.6 & 77.4 & 58.8 & 75.4 \\
         ClipCap* & 84.9 & 66.8 & 49.1 & 65.8\\ 
         \textsc{SmallCap} &87.6 & 78.6 & 68.9 & 77.9 \\
          \textsc{SmallCap}$_\text{{+W+H}}$ & \textbf{90.5} & \textbf{85.6} & \textbf{91.5} & \textbf{87.5} \\
     \bottomrule
     \end{tabular}
     }
     \caption{Validation results in CIDEr score on \texttt{nocaps}.  * Results copied from the respective publications. $\star$ Results computed by us. $\circ$ Results obtained through personal communication. }
     \label{tab:nocaps_val}
 \end{table}
In Table~\ref{tab:nocaps_val}, we show results on the \texttt{nocaps} validation set, since several recent studies only include performance on the validation set, following \cite{li2020oscar}. In line with the test set results, \textsc{SmallCap}$_{\text{+W+H}}$ outperforms other lightweight-training models and even the large model OSCAR, especially in the \textit{Out}-of-domain setting.

%\subsection{Analysis of the retrieved captions}
%After exploring the ability of \model to exploit large data in a training-free fashion in section \ref{varying_datastore}, we now discuss the setting \emph{In-domain + Web + Human-Labeled} in respect to the percentage of retrieved captions that come from each dataset. As can be seen in Figure \ref{fig:percentage}, most text is retrieved from Web, specifically in the presence of unseen visual concepts, such as \texttt{nocaps}. Moreover, there is a tendency to retrieve text from the corresponding dataset or from a similar domain, for instance, for MSR-VTT text is retrieved from video datasets. 

\section{\model with Alternative Decoders}
\label{app:scaling_dec}

%We study the  scaling behaviour of \model with larger decoders and a lower cross-attention dimensionality ($d=16$, $d=8$ and $d=4$), which allows us to leverage these larger decoders without a massive increase in the number of trainable parameters.\footnote{Notwithstanding, a larger decoder still requires more GPU memory, which means that we need to the reduce the batch size and use gradient accumulation to train \model$_{\text{Medium}}$ and \model$_{\text{Large}}$.} As seen in Figure \ref{fig:scale_fig}, the performance of \model${_\text{Large}}$ is extremely stable across the different dimensionalities of the cross-attention. 

\subsection{Larger GPT-2 decoders}

\begin{figure}[!h]
    \centering
    \includegraphics[width=\linewidth]{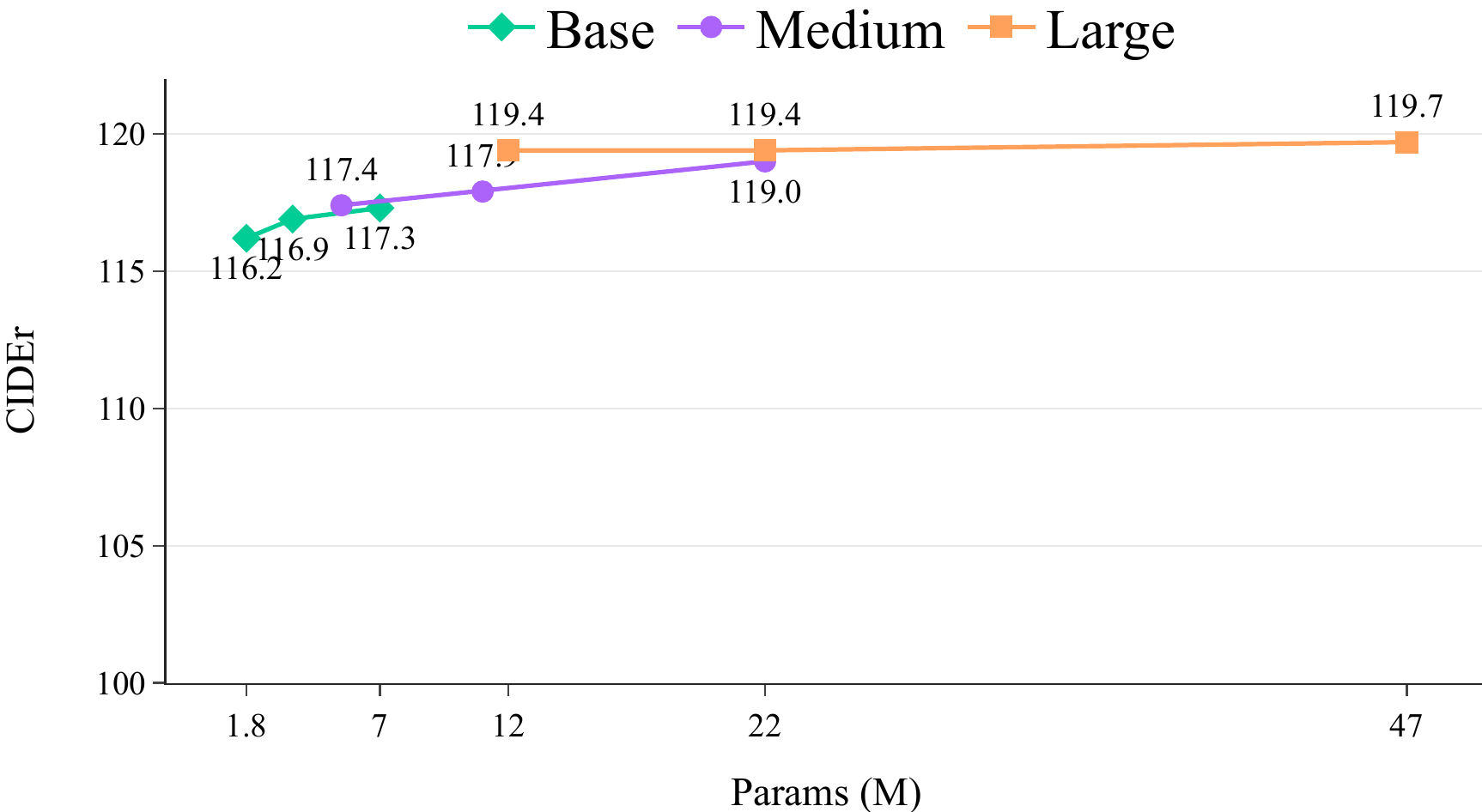}
    \caption{CIDEr performance on the COCO validation set across different decoder sizes: GPT-Base, GPT-Medium, GPT-Large and cross-attention dimensionalities  $d=4, 8, 16$ .}
    \label{fig:scale_fig}
\end{figure}

In Figure \ref{fig:scale_fig}, we study the  scaling behaviour of \model with larger decoders for different cross-attention dimensionalities ($d=4$, $d=8$ and $d=16$). We can see that it is beneficial to train with GPT-Medium
and GPT-Large across the different dimensionalities of the cross-attention. 
%This is similar to trend presented with \model$_{\text{Base}}$ in Figure \ref{} wherein performance keeps stable with lower number of trainable parameters. 
Controlling the cross-attention dimensionality allows us to leverage these larger decoders without a massive increase in the number of trainable parameters while maintaining a stable performance. Notwithstanding, a larger decoder still requires more GPU memory, which means that we had to reduce the batch size and use gradient accumulation to train \model$_{\text{Medium}}$ and \model$_{\text{Large}}$ models.

\begin{table}[]
    \centering
    \begin{tabular}{lll}
    \toprule
    & OPT-125M & OPT-350M \\
    \midrule
    With retrieval & 120.8 & 120.8   \\
    Without retrieval & 113.4 & 112.6\\
    \bottomrule
    \end{tabular}
    \caption{Validation results in CIDEr score on COCO.}
    \label{tab:opt-val}
\end{table}

\subsection{OPT decoders}

In Section~\ref{different dec}, we showed results for \model variants trained with a different decoder based on OPT \cite{zhang2022opt}. In Table~\ref{tab:opt-val} we report results from models trained with and without retrieval. The large drop in performance without retrieval demonstrates that retrieval is key to the good model performance with this different decoder, as was observed for \model using GPT-2 (see Section~\ref{cross_attention_reduction}). 

\section{Data}
\label{appendix_data}

For the experiments in Section~\ref{sec:training}, we explored different sources of data to include in the datastore, detailed in Table~\ref{tab:size_of_data}. Specifically, we used the cleaner web data version proposed in Li \etal~\cite{li2022blip}, which contains synthetic model-generated texts for the same web images, instead of using the original noisy web texts given the findings that noisy web texts are suboptimal for vision-and-language tasks \cite{li2022blip}. We also used different human-labeled data beyond image captioning datasets, including video captioning, audio captioning and localized narratives. We only included in the datastore text with length shorter than 25 tokens. Regarding the index, for human-labeled data, since it is limited-scale, we used \texttt{IndexFlatIP} without requiring training. For the web data, given its larger size, we used \texttt{IndexIVFFlat} with a training stage to speed up the search (with the hyperparameter \emph{nprobe} equal to 16). In terms of space, the COCO datastore takes up 2.2GB, the Human-Labeled datastore takes 8GB, and the Web datastore takes 49GB. Future work can include a further exploration of index types, since the FAISS library provides different indexes to customize for a faster search and lower memory footprint (e.g., through quantization). 

\begin{table}[]
    \centering
    \begin{tabular}{lrr}
        \toprule
         Dataset & Data type & Size \\ \midrule
        Web ~\cite{li2022blip}& Image captions & 12M  \\
        Human-Labeled &~ & 2.1M\\%115663\\
        ~~COCO  \cite{chen2015microsoft} & Image captions & 566K  \\
        ~~Flickr \cite{young2014image}& Image captions& 145K \\
        ~~VizWiz  \cite{gurari2020captioning}& Image captions&   117K\\
         ~~LN Ade20k\cite{pont2020connecting} & Image Narratives & 19K\\
        ~~LN COCO\cite{pont2020connecting}& Image Narratives& 121K \\
        ~~LN Flick30k\cite{pont2020connecting} & Image Narratives& 28K   \\
        ~~LN Open Images \cite{pont2020connecting}& Image Narratives &496K\\%,472\\
        ~~MSR-VTT  \cite{xu2016msr}& Video captions &  130K \\
        ~~VATEX  \cite{wang2019vatex}& Video captions & 349K  \\
        ~~TGIF    \cite{li2016tgif} & GIF captions & 125K  \\
        ~~Clotho  \cite{drossos2020clotho}& Audio captions& 14K  \\
        \bottomrule
    \end{tabular}
    \caption{Data used in the datastore for the experiments reported in Section~\ref{datastore_explained} along with size in terms of image-caption pairs. LN stands for Localized Narratives \cite{pont2020connecting}.} 
    \label{tab:size_of_data}
\end{table}

\section{Inference time}
\label{inference_time}

\model is a lightweight-training captioning model. Although training efficiency is of crucial importance, especially in contexts involving limited resources, inference time should also be to taken into account. We thus measured the inference time of \model and CaMEL on an NVIDIA A100 GPU across 1,000 randomly sampled images from COCO. The resulting values are 0.22 and 0.58 seconds per image, respectively, i.e., \model is much faster than CaMEL, likely due to CaMEL's dual decoder architecture. In Section~\ref{experimental_setup}, we also report the residual difference of generating a caption with and without retrieval at inference time.

\section{More Qualitative Examples}
\label{examplas_without_retrieval}

%for the Flickr30k image \model is less biased towards the very frequent concept \textit{horse} compared to the correct concept, \textit{camel}; for the VizWiz image it is more descriptive and uses the Swanson brand name,  never observed in the COCO training data, and for the MSR-VTT image it refers to Pokemon, a concept observed just six times in the COCO training data. } 

%In Figure~\ref{fig:with_and_without_retrieval}, we provide examples of generated captions on the COCO dataset by \model compared to its variant trained without retrieval. 

\begin{figure}[!h]
    \centering
    \includegraphics[width=0.9\linewidth]{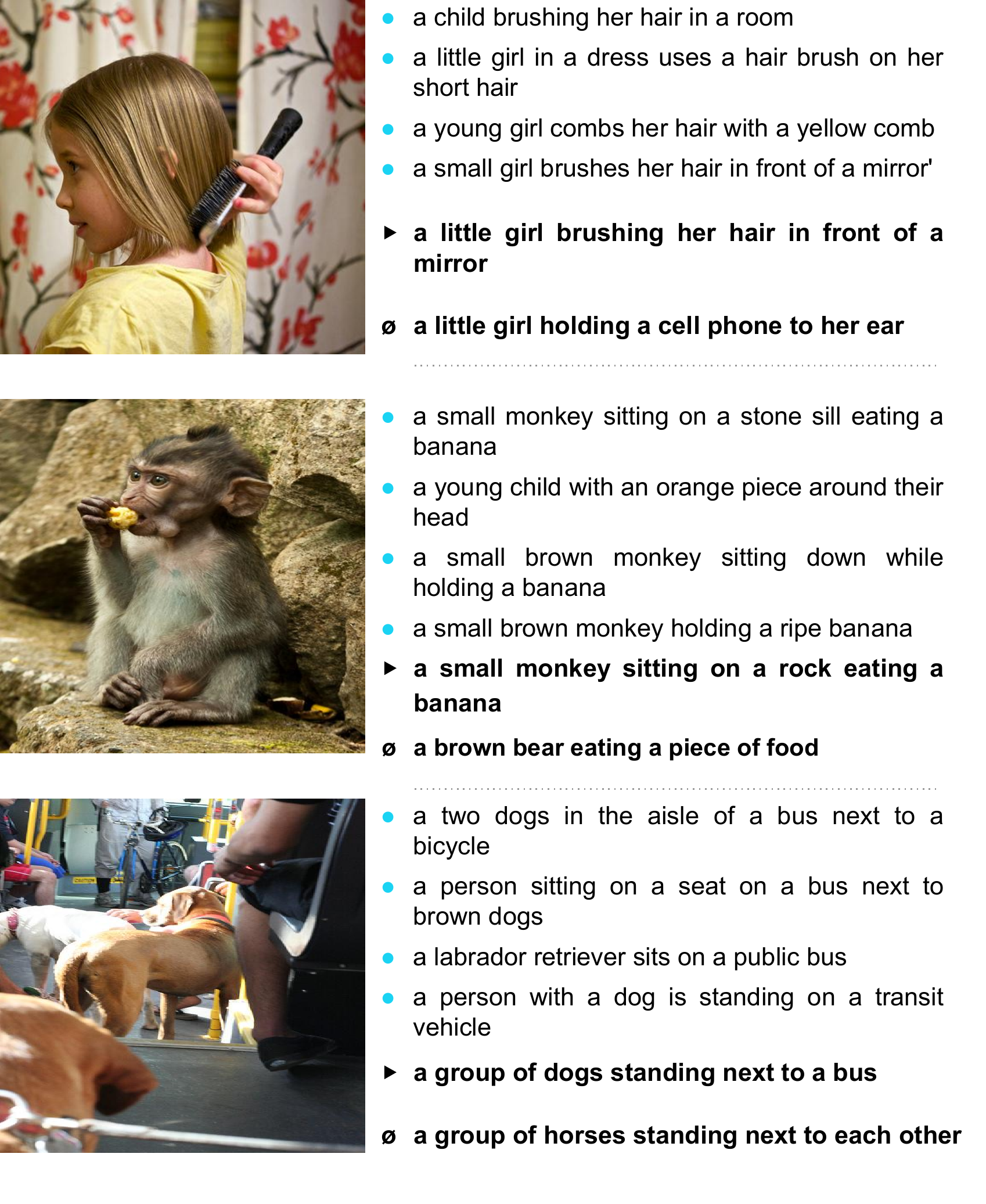}
    \caption{Caption examples from COCO generated with and without retrieval augmentation. \cococolor{$\bullet$} denotes the retrieved captions, $\blacktriangleright$ denotes the generated caption from \model; $\o$ denotes the caption generated by a model trained without retrieval augmentation. }
    \label{fig:with_and_without_retrieval}
\end{figure}

Figure~\ref{fig:with_and_without_retrieval} shows examples of captions generated by \model on the COCO dataset, compared to a variant trained without retrieval. In line with the quantitative results that were presented before, \model can better describe an input image when conditioning on the retrieved examples. In the first picture, we see that without retrieval a brush is mistaken for a cell phone, which is a more common object in the COCO training data.

\begin{figure*}[!h]
    \centering
    \includegraphics[width=1.0\textwidth]{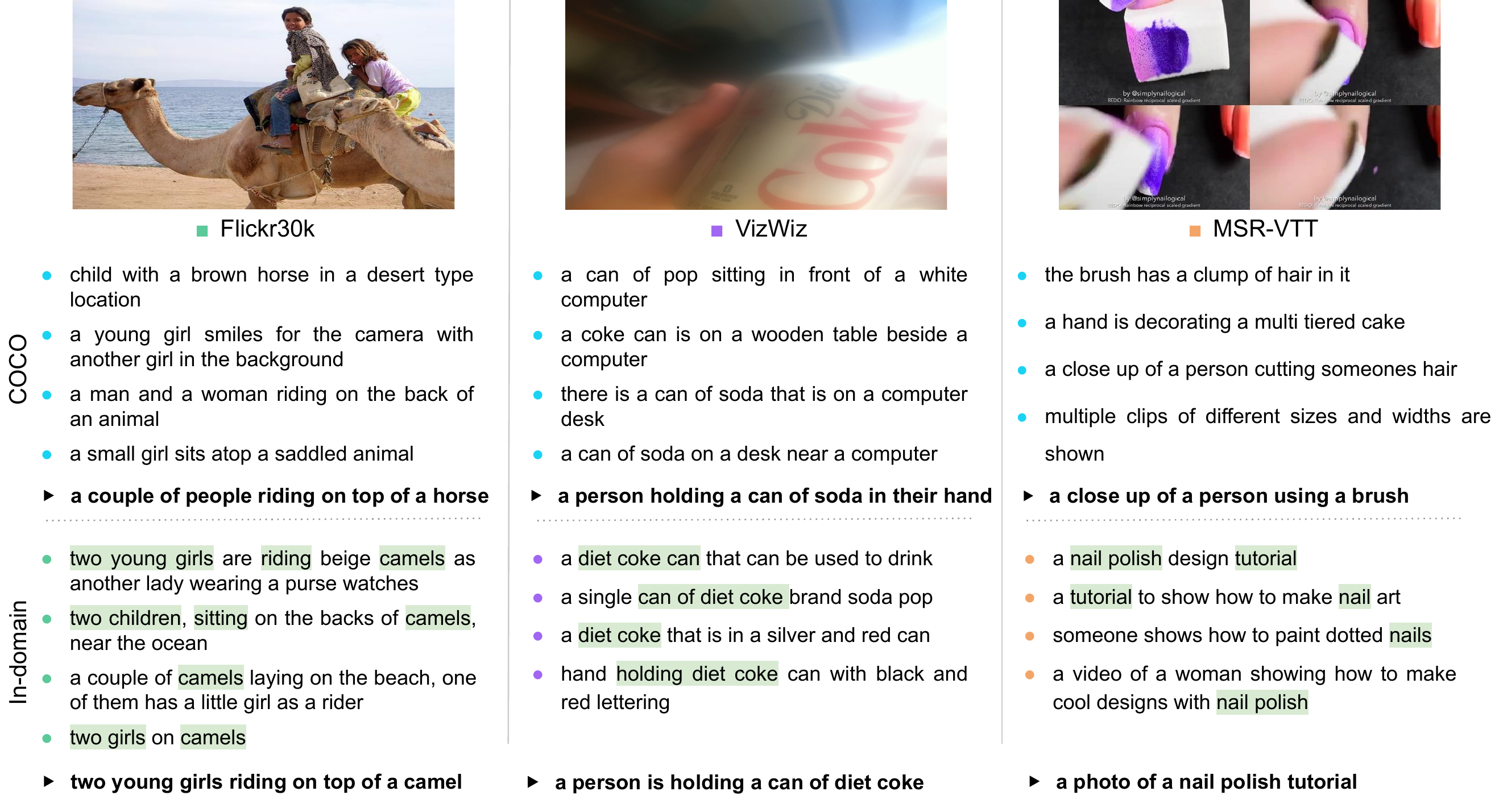}
    \caption{Captions generated for images from the Flickr30k, VizWiz, and MSR-VTT datasets, with retrieval either from COCO or in-domain data. With retrieval from in-domain data, \model is less biased towards very frequent concepts, such as \textit{horse}, \textit{soda}, or \textit{brush}, compared to the correct concepts, respectively \textit{camel}, \textit{diet coke} and \textit{nail polish}.}
    \label{fig:domain2}
\end{figure*}

In addition, in Figure~\ref{fig:domain2}, we provide more examples of captions illustrating how \model adapts to Flickr30k, VizWiz, and MSR-VTT, by replacing the contents of the datastore with the in-domain data.

Lastly, we measured the importance of generating a caption conditioned on retrieved information compared to directly using the nearest caption as the prediction (i.e., image captioning through retrieval alone). The latter approach yields a CIDEr score of 65.5 on the COCO validation set, substantially lower than the 117.3 from \model.

%\subsection{Analysis}

%Look at breakdown between retrieval from web v labeled when we have both

%Look at the breakdown between retrieval from the different human-labeled datasets in experiments with labeled data (decide if with in-domain or without)

\end{document}